%% file: main.tex
\documentclass[lettersize,journal]{IEEEtran}
\usepackage{amsmath,amsfonts,amssymb}
\usepackage{algorithmic}
\usepackage{algorithm}
\usepackage{array}
\usepackage[caption=false,font=normalsize,labelfont=sf,textfont=sf]{subfig}
\usepackage{textcomp}
\usepackage{stfloats}
\usepackage{url}
\usepackage{verbatim}
\usepackage{graphicx}
\usepackage{cite}
\usepackage{booktabs}
\usepackage{multirow}
\usepackage{xcolor}
\usepackage{bm}
\definecolor{rankfirst}{HTML}{FF0000}
\definecolor{ranksecond}{HTML}{0000FF}
\definecolor{rankthird}{HTML}{008000}

\makeatletter
\long\def\@IEEEtitleabstractindextextbox#1{\parbox{0.922\textwidth}{#1}}
\makeatother

\begin{document}

\title{Semantic-Aware Temporal Adaptation for UAV Anti-UAV Tracking}

\author{Xiaozhen~Qiao,
        Da~Zhang,
        Yubin~Guo,
        Junyu Gao,
        Zhiyuan~Zhao,
        and~Xuelong~Li,~\IEEEmembership{Fellow,~IEEE}
\thanks{Manuscript received XXXX XX, XXXX; revised XXXX XX, XXXX.}
\thanks{X. Qiao is with the School of Information Science and Technology, University of Science and Technology of China, Hefei 230026, China. This work was done during a research internship at the Institute of Artificial Intelligence (TeleAI), China Telecom, P. R. China (e-mail: xiaozhennnqiao@mail.ustc.edu.cn).}
\thanks{D. Zhang and J. Gao are with Northwestern Polytechnical University, Xi'an 710072, China, and also with the Institute of Artificial Intelligence (TeleAI), China Telecom, P. R. China.}
\thanks{Y. Guo is with the University of Science and Technology of China, Hefei, China, and also with the Institute of Artificial Intelligence (TeleAI), China Telecom, P. R. China.}
\thanks{Z. Zhao and X. Li are with the Institute of Artificial Intelligence (TeleAI), China Telecom, P. R. China.}
\thanks{Corresponding author: Xuelong Li (e-mail: xuelong\_li@ieee.org).}}

\markboth{Journal of \LaTeX\ Class Files,~Vol.~14, No.~8, August~2021}%
{Qiao \MakeLowercase{\textit{et al.}}: Semantic-Aware Temporal Adaptation for UAV Anti-UAV Tracking}

\IEEEtitleabstractindextext{%
\input{section/0_abstract}}

\maketitle

\IEEEdisplaynontitleabstractindextext

\IEEEpeerreviewmaketitle

\section{Introduction}
\input{section/1_intro}

\section{Related Work}
\input{section/2_related}

\section{Method}

\input{section/3_methods}

\section{Experiments}
\input{section/4_exp}

\section{Discussion}
\input{section/5_discussion}

\section{Conclusion}
\input{section/6_conclusion}

\enlargethispage{10pt}
\bibliographystyle{IEEEtran}
\bibliography{reference}



\end{document}

%% file: section/0_abstract.tex
\begin{abstract}
UAV Anti-UAV tracking is an emerging low-altitude security task for localizing an adversarial UAV using the onboard camera of a moving observer UAV. It differs from conventional UAV tracking and ground-based Anti-UAV tracking because both the camera platform and the target move simultaneously. This dual-dynamic setting induces rapid viewpoint changes, motion blur, scale variation, and visually similar distractors, making reliable appearance matching difficult. Under such rapidly changing conditions, fixed visual representations are often insufficient because target appearance becomes unreliable and feature distributions may deviate from the training domain. The target language description remains stable across frames and can therefore serve as a semantic anchor for temporal state propagation, while online feature-distribution alignment can reduce video-specific test-time shifts. In this paper, we propose \emph{SATATrack}, a Semantic-Aware Temporal Adaptation framework for UAV Anti-UAV tracking. SATATrack introduces Semantic-Aware Context Propagation (SACP), which uses the target description to guide temporal context propagation across backbone stages and preserve target identity under rapid appearance changes. An auxiliary contrastive regularizer is used during training to discourage responses to semantically similar background regions. During inference, Temporal-Aware Distribution Alignment (TADA) aligns feature distributions online without updating model parameters, combining recent-frame estimates with training-time statistics for stability. SATATrack achieves state-of-the-art performance on the UAV-Anti-UAV benchmark while remaining competitive in Anti-UAV and UAV object tracking tasks. The code will be available at \url{https://github.com/XiaozhenQiao/SATATrack}.
\end{abstract}

\begin{IEEEkeywords}
UAV Anti-UAV tracking, Vision-language tracking, Temporal adaptation, State-space model.
\end{IEEEkeywords}

%% file: section/1_intro.tex
\IEEEPARstart{U}{AV} Anti-UAV tracking is an important problem for low-altitude security and AI-driven aerial perception~\cite{an2026aiflow,chen2026gvc}, where a moving observer UAV must localize an adversarial UAV during air-to-air motion~\cite{zhang2025far}. It differs from conventional UAV object tracking, which mainly studies UAV-mounted cameras observing ground objects such as vehicles or pedestrians~\cite{mueller2016benchmark,du2018unmanned,zhu2020vision,zhang2022webuav}, and from ground-based Anti-UAV tracking, which typically observes flying UAVs with static or limited-motion sensors~\cite{jiang2021anti,zhao2022vision,huang2023anti,zhu2023evidential,xu2025tri}. In UAV Anti-UAV tracking, both the camera platform and the target move simultaneously, resulting in severe dual-dynamic disturbance, rapid viewpoint transition, background turbulence, motion blur, and abrupt scale variation. The target UAV is often tiny, texture-poor, and visually similar to background objects or other flying distractors, making reliable appearance matching difficult.

\begin{figure*}[t]
    \centering
    \includegraphics[width=1.0\textwidth]{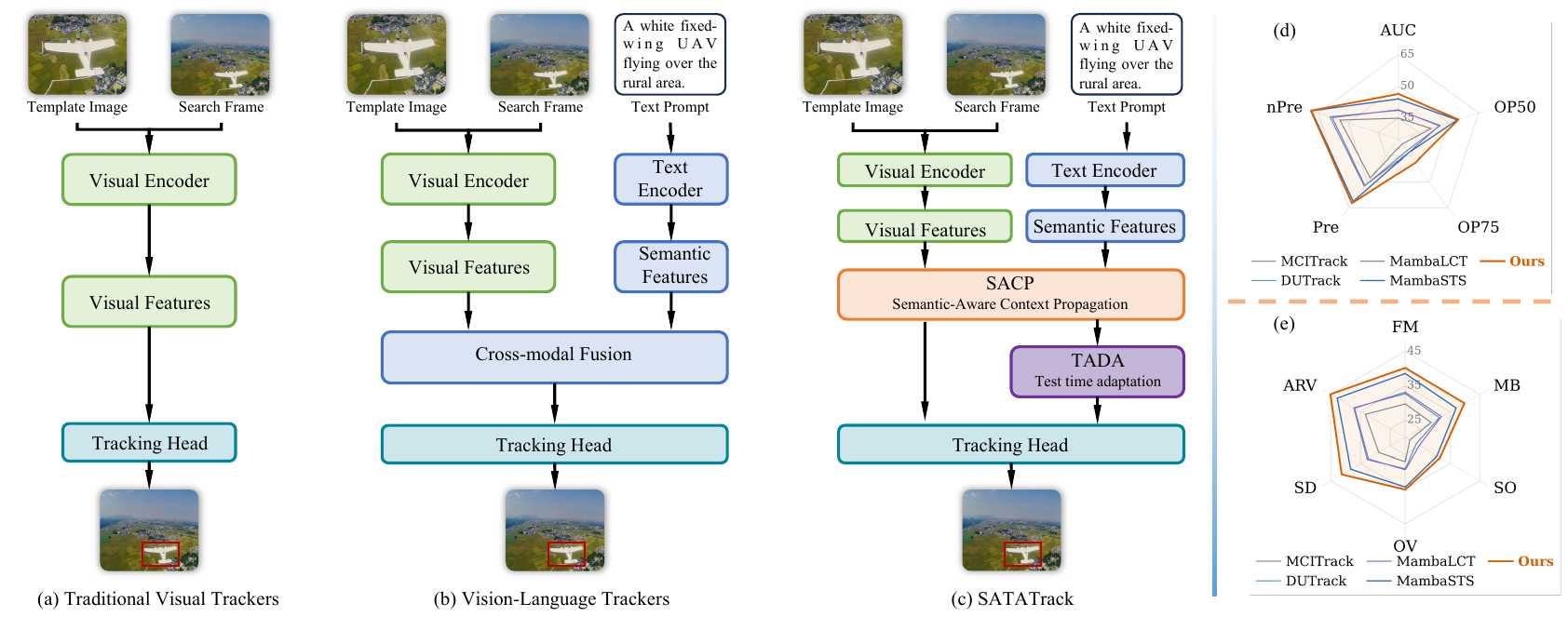}
    \caption{(a) Traditional visual trackers. (b) Existing vision-language trackers. (c) The proposed SATATrack, which adds SACP and TADA. (d) Radar of the five UAV-Anti-UAV metrics. (e) Radar of attribute-wise AUC on UAV-Anti-UAV.}
    \label{fig:teaser}
\end{figure*}

Existing trackers are not well matched to this air-to-air tracking problem. When a UAV appears small or blurred among flying distractors, a tracker must match the current frame while deciding which past evidence remains trustworthy. Pure visual trackers with strong convolutional or Transformer backbones~\cite{bertinetto2016fully,danelljan2019atom,li2019siamrpn++,ye2022joint,zheng2024odtrack} can accumulate wrong responses once appearance cues become ambiguous. Vision-language trackers bring useful semantic priors~\cite{wang2021towards,guo2022divert,zhou2023joint,li2023citetracker,zhang2023all,ma2024unifying,chen2025sutrack,li2025dynamic,feng2025atctrack}, but the language cue is mostly consumed at the matching or query stage. Recent state-space trackers improve efficient temporal propagation~\cite{gu2023mamba,zhu2024vision,wu2024mambanut,li2025mambalct,kang2025exploring,zhang2025mambatrack}, and MambaSTS~\cite{zhang2025far} further provides a strong UAV-Anti-UAV baseline with spatial-temporal-semantic learning. Yet even these models fuse semantics as an extra feature while the state transition itself stays driven by visual observations, leaving propagated states vulnerable to distractor contamination. Another difficulty arises from test-time feature shifts, since fixed normalization statistics from generic training data may mismatch video-specific aerial distributions, while online parameter updates can amplify errors from unreliable predictions~\cite{qiao2026bidirectional,qiao2025class}. These observations suggest two requirements: temporal memory should be conditioned on target identity, and test-time adaptation should be performed without modifying learned weights.

Building on this motivation, we propose \emph{SATATrack}, a Semantic-Aware Temporal Adaptation framework for UAV Anti-UAV tracking. SATATrack first introduces Semantic-Aware Context Propagation (SACP), which uses the target description to guide state-space context propagation across backbone stages. Specifically, SACP injects the target semantic embedding into the temporal update so that the propagated state can emphasize target-relevant evidence and suppress distractor cues when visual appearance changes abruptly. The updated context is further injected into the visual stream at multiple feature levels, providing persistent semantic guidance rather than a one-shot language cue. An auxiliary contrastive regularizer discourages responses to semantically similar background regions during training. During inference, Temporal-Aware Distribution Alignment (TADA) aligns feature distributions online without updating model parameters. It combines recent-frame estimates with training-time statistics, allowing the tracker to adapt to each test video while preserving the discriminative knowledge learned offline. The resulting tracker targets both identity drift and video-specific feature shifts in dynamic UAV-to-UAV tracking.

We conduct our main evaluation using the UAV-Anti-UAV benchmark~\cite{zhang2025far}. We also report results on Anti-UAV318~\cite{jiang2021anti}, DUT Anti-UAV~\cite{zhao2022vision}, UAV123~\cite{mueller2016benchmark}, and UAVDT~\cite{du2018unmanned} to examine whether the tracker remains robust in related Anti-UAV and UAV object tracking tasks. SATATrack achieves state-of-the-art performance on UAV-Anti-UAV and competitive results on the additional benchmarks, as previewed in Fig.~\ref{fig:teaser}. The main contributions of this work are summarized below.
\begin{itemize}
    \item We propose SATATrack, a vision-language tracker for UAV Anti-UAV tracking that combines target-language-guided temporal modeling with inference-time feature-distribution alignment.
    \item We design SACP, which conditions the state-space temporal update on the target language, letting semantics control what the propagated memory retains rather than serving as a one-shot matching cue.
    \item We introduce TADA, a lightweight inference-time adaptation scheme that aligns feature distributions online without updating model parameters.
    \item We conduct comprehensive experiments centered on UAV-Anti-UAV, showing state-of-the-art performance on this benchmark and competitive results on related Anti-UAV and UAV object tracking tasks.
\end{itemize}

%% file: section/2_related.tex
\subsection{UAV and Anti-UAV Tracking}

Generic single-object tracking has progressed rapidly with large-scale benchmarks and stronger matching architectures. Benchmarks such as GOT-10k~\cite{huang2019got}, LaSOT~\cite{fan2019lasot}, and TrackingNet~\cite{muller2018trackingnet} provide diverse training and evaluation sources for learning category-agnostic target representations. Based on these data, Siamese trackers~\cite{bertinetto2016fully,li2019siamrpn++}, discriminative online trackers~\cite{danelljan2017eco,danelljan2019atom}, and Transformer-based trackers~\cite{chen2021transformer,ye2022joint,zheng2024odtrack} have improved appearance matching, target-background discrimination, and temporal relation modeling. UAV tracking further stresses these models with low-altitude camera motion, viewpoint change, scale variation, and motion blur. Aerial benchmarks such as UAV123~\cite{mueller2016benchmark}, UAVDT~\cite{du2018unmanned}, VisDrone~\cite{zhu2020vision}, and WebUAV-3M~\cite{zhang2022webuav} study generic objects captured by UAV-mounted cameras, while aerial trackers such as HiFT~\cite{cao2021hift}, TCTrack~\cite{cao2022tctrack}, adaptive background-aware tracking~\cite{li2023adaptive}, SLA-ViT~\cite{xue2025similarity}, and occlusion-robust UAV tracking~\cite{wu2025learning} exploit hierarchical features, temporal contexts, background modeling, and efficient Transformer designs for aerial scenarios. Anti-UAV tracking differs from generic UAV object tracking because the target is itself an unauthorized UAV, often appearing as a tiny, texture-poor object against cluttered sky, building, or vegetation backgrounds. Anti-UAV benchmarks such as Anti-UAV318~\cite{jiang2021anti}, DUT Anti-UAV~\cite{zhao2022vision}, Anti-UAV410~\cite{huang2023anti}, Anti-UAV600~\cite{zhu2023evidential}, and MM-UAV~\cite{xu2025tri} introduce visible, infrared, and multi-modal settings for drone detection and tracking, covering target disappearance, low contrast, thermal ambiguity, and long-term re-localization. Recent surveys also highlight the importance of robust benchmarking for practical counter-UAV systems~\cite{dong2025securing}. However, most existing Anti-UAV datasets are captured by static or limited-motion ground sensors. The UAV-Anti-UAV benchmark~\cite{zhang2025far} moves one step further by introducing an air-to-air setting, where both the observer UAV and the target UAV move simultaneously. This dual-dynamic configuration causes severe relative motion, abrupt background changes, frequent scale variation, and stronger target-background ambiguity, making direct transfer from either generic UAV tracking or ground-based Anti-UAV tracking insufficient.

\subsection{Vision-Language Tracking}

Vision-language tracking introduces semantic descriptions to complement visual appearance, providing target-level priors when purely visual evidence becomes ambiguous. Early language-guided tracking studies~\cite{wang2021towards} show that natural language can make target specification more flexible than box-only initialization, especially when multiple visually similar objects exist in the search region. Large-scale language and vision-language models such as BERT~\cite{devlin2018bert} and CLIP~\cite{radford2021learning} further improve the quality of textual and cross-modal representations. Building on these representations, VLT$_{\rm TT}$~\cite{guo2022divert} diverts attention to language-relevant visual regions, JointNLT~\cite{zhou2023joint} studies joint visual grounding and tracking, CiteTracker~\cite{li2023citetracker} correlates image and text cues for target localization, and All-in-One~\cite{zhang2023all} and UVLTrack~\cite{ma2024unifying} pursue unified formulations across visual and vision-language tracking. Recent methods further explore how language should interact with tracking dynamics. SUTrack~\cite{chen2025sutrack} simplifies unified tracking with multi-source inputs, DUTrack~\cite{li2025dynamic} studies dynamic language adaptation, and ATCTrack~\cite{feng2025atctrack} aligns target-context cues with dynamic target states. These methods demonstrate that semantic cues can improve robustness when appearance alone is unreliable. Nevertheless, most existing vision-language trackers utilize language primarily at the matching stage for cross-modal alignment, query enhancement, or feature fusion, without explicitly leveraging it to guide the temporal evolution of the tracker. This limitation is particularly critical in UAV anti-UAV tracking, where target descriptions remain stable across frames, whereas visual observations often suffer from severe scale reduction, motion blur, or background confusion from UAV-like structures. Consequently, language information is most informative when maintained as a persistent temporal cue throughout tracking.

\subsection{Temporal Modeling and Adaptation}

Temporal modeling is essential for UAV anti-UAV tracking, as single-frame observations often provide insufficient evidence for stable localization. Classical trackers address this via online appearance updating or discriminative optimization, including ECO~\cite{danelljan2017eco}, ATOM~\cite{danelljan2019atom}, and Siamese matching frameworks~\cite{bertinetto2016fully}. Transformer-based methods improve temporal reasoning through global attention~\cite{dosovitskiy2020image,chen2021transformer}, while temporal Transformers such as TrDiMP~\cite{wang2021transformer}, STARK~\cite{yan2021learning}, TCTrack~\cite{cao2022tctrack}, and ODTrack~\cite{zheng2024odtrack} further aggregate cross-frame context for robust tracking. However, their temporal memory is mainly driven by visual similarity and prediction confidence, which can be unreliable under abrupt motion, scale variation, or UAV-like distractors, leading to drift. State-space models offer an efficient alternative for long-sequence modeling. Mamba~\cite{gu2023mamba} and Vision Mamba~\cite{zhu2024vision} introduce selective state-space dynamics for linear-time sequence modeling with input-dependent transitions, and recent tracking methods including MambaNUT~\cite{wu2024mambanut}, MambaLCT~\cite{li2025mambalct}, MCITrack~\cite{kang2025exploring}, and MambaTrack~\cite{zhang2025mambatrack} demonstrate their effectiveness in UAV and video tracking. In particular, MambaSTS~\cite{zhang2025far} integrates spatial, temporal, and semantic modeling within a Mamba framework, but existing semantic SSM trackers still treat language as an auxiliary cue rather than using it to directly steer state transitions. SATATrack addresses this limitation by conditioning state-space updates on target descriptions, enabling semantic guidance to influence temporal propagation across the visual hierarchy. Beyond temporal modeling, aerial tracking also suffers from significant domain shift due to variations in sensor characteristics, illumination, altitude, scale, and background statistics. Direct online adaptation is often unstable, as unreliable predictions may accumulate and amplify drift, especially for small or intermittently occluded targets. To mitigate this, SATATrack adopts a conservative test-time adaptation strategy that recalibrates feature statistics using recent observations while keeping model parameters fixed, complementing semantic-aware propagation by preserving target identity while reducing feature distribution mismatch without gradient-based updates.

\begin{figure*}[!t]
    \centering
    \includegraphics[width=\textwidth]{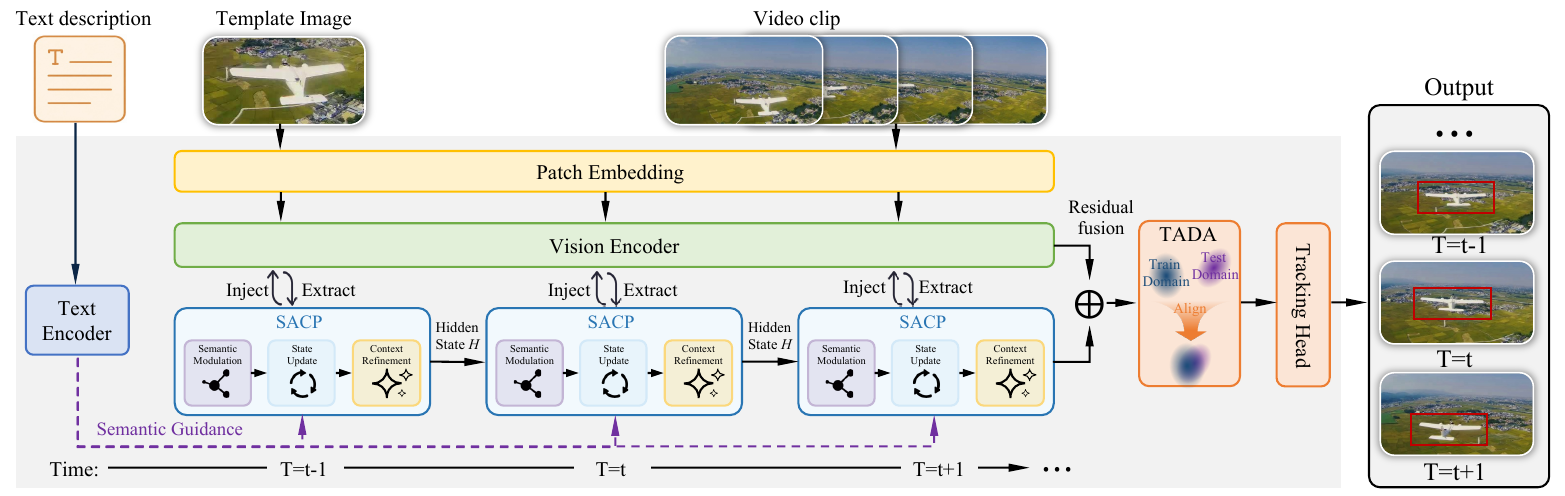}
    \caption{Overview of the proposed pipeline. Given a template image, a search image, and a semantic description of the target, the framework extracts multi-level visual features with the backbone, propagates target-aware temporal context across stages, and predicts the target location with the localization head.}
    \label{fig:framework}
\end{figure*}

%% file: section/3_methods.tex
In this section, we present the proposed framework in detail. We first provide an overview of the overall pipeline, and then describe the backbone, the semantic-aware context propagation module, and the loss function used during training, including semantic-discriminative contrastive learning. Finally, we introduce the temporal-aware distribution alignment module that operates purely at test time to bridge the domain gap between training and inference.

\subsection{Overview}
\label{sec:overview}

The overall architecture of the proposed framework is illustrated in Fig.~\ref{fig:framework}. Given a template image $\mathbf{Z} \in \mathbb{R}^{H_z \times W_z \times 3}$, a search image $\mathbf{X}_t \in \mathbb{R}^{H_x \times W_x \times 3}$ at time step $t$, and a semantic description of the target, our goal is to accurately localize the UAV in the search region. The framework consists of three components: a backbone for visual feature extraction, a semantic-aware context propagation module for temporal modeling, and a localization head. During training, semantic-discriminative contrastive learning further sharpens the separation between the target UAV and confusing background regions. At test time, a temporal-aware distribution alignment module recalibrates the feature statistics on the fly to bridge the gap between the training and inference domains, without updating any learned parameters. Specifically, the template and search images are first divided into non-overlapping patches and fed into a shared backbone network. Let $\mathbf{F}_t^{i-1}$ denote the visual tokens before the $i$-th backbone block, and let $\mathbf{C}_t^{i-1}$ denote the propagated context tokens at the same stage. The semantic description is encoded into a global semantic embedding, which is used to modulate the state-space update in the propagation branch. At each stage, the semantic-aware context propagation module first updates the propagated context with historical hidden states, then injects the updated context into the visual stream, and finally delivers the refined features to the localization head.

\subsection{Backbone}
We adopt Fast-iTPN as the visual backbone because its narrow-and-deep architecture provides multi-level representations that are well suited to stage-wise context propagation. The template image $\mathbf{Z}$ and the search image $\mathbf{X}_t$ are first tokenized into patch embeddings and then processed by a stack of backbone blocks. For the $i$-th block, we denote the input visual tokens by $\mathbf{F}_t^{i-1} \in \mathbb{R}^{K \times d}$ and the output by $\mathbf{F}_t^{i} \in \mathbb{R}^{K \times d}$. Each backbone stage is associated with a propagation branch, enabling temporal context to be injected at multiple feature levels rather than only at the final layer. In addition, we employ a CLIP text encoder to transform the target description into a global semantic embedding $\mathbf{s} \in \mathbb{R}^{d_s}$, which serves as the guidance signal for the subsequent context propagation module. Therefore, the framework leverages both appearance cues from Fast-iTPN and semantic cues from CLIP for anti-UAV tracking.

\subsection{Semantic-Aware Context Propagation}
\label{sec:context}
\begin{figure}[!t]
    \centering
    \includegraphics[width=\columnwidth]{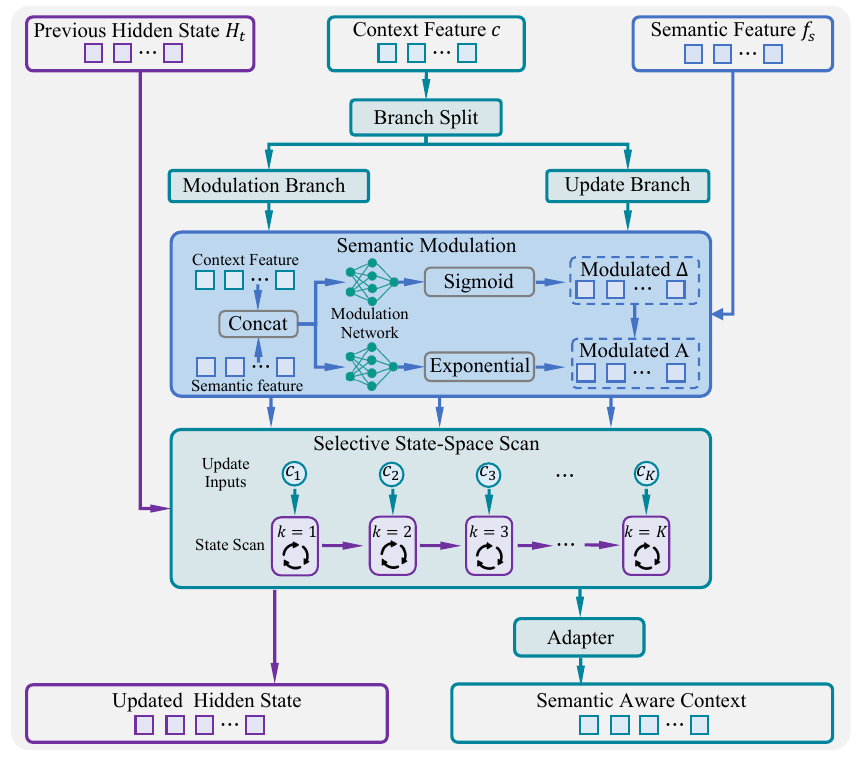}
    \caption{Pipeline of the proposed SACP module, where the target semantics modulate the state-space update to produce target-aware temporal context.}
    \label{fig:sacp}
\end{figure}

To model long-range temporal dependencies while suppressing distractors, we introduce the SACP module, illustrated in Fig.~\ref{fig:sacp}. Unlike conventional SSM-based propagation~\cite{gu2023mamba}, which updates hidden states solely according to visual observations, SACP incorporates target semantics into the state update process, thereby producing target-aware temporal context.

As shown in Fig.~\ref{fig:sacp}, each propagation stage proceeds in three steps: a semantic-aware state-space update of the propagated context, injection of the updated context into the visual stream, and a write-back of the current visual evidence to the propagation branch. For the $i$-th propagation stage, let $\mathbf{F}_t^{i-1} \in \mathbb{R}^{K \times d}$ be the visual tokens from the backbone and $\mathbf{C}_t^{i-1} \in \mathbb{R}^{K \times d}$ be the context tokens. Given the hidden state $\mathbf{H}_{t-1}^{i}$ from the previous frame, SACP first applies a semantic-aware state-space update:
\begin{equation}
    \bar{\mathbf{C}}_t^{i}, \mathbf{H}_t^{i} = \operatorname{SACP}_{i}(\mathbf{C}_t^{i-1}, \mathbf{H}_{t-1}^{i}, \mathbf{s}),
\end{equation}
where $\mathbf{s}$ denotes the semantic embedding of the target and the internal operations of $\operatorname{SACP}_{i}$ are detailed from Eq.~(\ref{eq:sacp_proj}) to Eq.~(\ref{eq:sacp_out}). The updated context is then injected into the visual stream:
\begin{equation}
    \tilde{\mathbf{F}}_t^{i-1} = \mathbf{F}_t^{i-1} + \operatorname{Attn}_{\mathrm{in}}^{i}(\mathbf{F}_t^{i-1}, \bar{\mathbf{C}}_t^{i}, \bar{\mathbf{C}}_t^{i}),
\end{equation}
\begin{equation}
    \mathbf{F}_t^{i} = \operatorname{Block}_{i}(\tilde{\mathbf{F}}_t^{i-1}).
\end{equation}
The current visual evidence is then written back to the propagation branch:
\begin{equation}
    \mathbf{C}_t^{i} = \operatorname{Attn}_{\mathrm{out}}^{i}(\bar{\mathbf{C}}_t^{i}, \mathbf{F}_t^{i}, \mathbf{F}_t^{i}) + \bar{\mathbf{C}}_t^{i}.
\end{equation}
By repeating the above process across backbone stages and video frames, SACP continuously propagates target-aware temporal context and provides more reliable guidance for UAV localization. We now detail the internal $\operatorname{SACP}_{i}$ operator, where semantic cues are injected directly into the selective state-space dynamics. Given the input context sequence $\mathbf{C}_t^{i-1}$, we first project it into two branches:
\begin{equation}
    [\mathbf{U}_t^{i}, \mathbf{V}_t^{i}] = \mathbf{W}_{in}\mathbf{C}_t^{i-1},
    \label{eq:sacp_proj}
\end{equation}
where $\mathbf{U}_t^{i}, \mathbf{V}_t^{i} \in \mathbb{R}^{K \times d_m}$ denote the state-update branch and the output-modulation branch, respectively. Meanwhile, the semantic embedding is projected into the SSM hidden space:
\begin{equation}
    \mathbf{s}' = \mathbf{W}_s \mathbf{s} + \mathbf{b}_s,
\end{equation}
where $\mathbf{W}_s \in \mathbb{R}^{d_m \times d_s}$ and $\mathbf{b}_s \in \mathbb{R}^{d_m}$ are learnable parameters.

The key design question is \emph{where} target semantics should enter this recurrence. In a selective state-space model, the hidden state $\mathbf{h}_{t,k}^{i}$ serves as a temporal memory, and two quantities govern how this memory evolves: the input-dependent step size $\boldsymbol{\Delta}$ and the state-transition matrix $\mathbf{A}^{i}$, which together form the discrete transition $\bar{\mathbf{A}}=\exp(\boldsymbol{\Delta}\mathbf{A}^{i})$ and the input gain $\bar{\mathbf{B}}=\boldsymbol{\Delta}\mathbf{B}$. Their roles are complementary. The step size $\boldsymbol{\Delta}$ behaves as a \emph{write/forget gate}: as $\boldsymbol{\Delta}\!\to\!\mathbf{0}$ we have $\bar{\mathbf{A}}\!\to\!\mathbf{I}$ and $\bar{\mathbf{B}}\!\to\!\mathbf{0}$, so the current observation is ignored and the existing memory is preserved, whereas as $\boldsymbol{\Delta}$ grows we have $\bar{\mathbf{A}}\!\to\!\mathbf{0}$ and the state is overwritten by the current token. The matrix $\mathbf{A}^{i}$ instead sets the \emph{retention timescale}, i.e., how slowly accumulated context decays once it has been written. We therefore inject semantics precisely into these two controls rather than fusing it as an auxiliary feature: the target description decides \emph{which} observations are allowed to write into the temporal memory (through $\boldsymbol{\Delta}$) and \emph{how long} target-consistent context is retained (through $\mathbf{A}^{i}$). Under this view, the target language is no longer a per-frame matching cue but a direct controller of the temporal memory itself, a control locus left untouched by prior vision-language trackers.

Concretely, for each token $\mathbf{u}_{t,k}^{i} \in \mathbb{R}^{d_m}$ in $\mathbf{U}_t^{i}$, the selective scan first computes a data-dependent step size
\begin{equation}
    \boldsymbol{\Delta}_{t,k}^{i} = \operatorname{Softplus}(\mathbf{W}_{\Delta}\mathbf{u}_{t,k}^{i} + \mathbf{b}_{\Delta}).
\end{equation}

Semantic conditioning is injected into two complementary components with distinct temporal roles. At the token level, we modulate the write strength by a semantic gate conditioned on the projected embedding $\mathbf{s}'$,
\begin{equation}
\begin{aligned}
\mathbf{g}_{t,k}^{i} &= \sigma\big(\operatorname{MLP}([\mathbf{u}_{t,k}^{i} ; \mathbf{s}'])\big), \\
\tilde{\boldsymbol{\Delta}}_{t,k}^{i} &= \boldsymbol{\Delta}_{t,k}^{i} \odot (1 + \eta\, \mathbf{g}_{t,k}^{i}),
\end{aligned}
\end{equation}
which adaptively amplifies tokens consistent with the target semantics while suppressing distractors.

At the sequence level, we modulate the retention dynamics by conditioning the log-parameterized transition matrix on the stable target description,
\begin{equation}
    \tilde{\mathbf{A}}^{i} = -\exp\!\big(\mathbf{a}^{i} + \mathbf{W}_{A}\mathbf{s} + \mathbf{b}_{A}\big),
\end{equation}
where the additive formulation preserves the stability of the exponential parameterization while enabling semantic control over channel-wise decay rates.

The remaining recurrence then follows the standard discretized state-space update with the semantically modulated transition,
\begin{equation}
    \bar{\mathbf{A}}_{t,k}^{i} = \exp(\tilde{\boldsymbol{\Delta}}_{t,k}^{i} \tilde{\mathbf{A}}^{i}), \qquad
    \bar{\mathbf{B}}_{t,k}^{i} = \tilde{\boldsymbol{\Delta}}_{t,k}^{i} \mathbf{B}_{t,k}^{i},
\end{equation}
\begin{equation}
    \mathbf{h}_{t,k}^{i} = \bar{\mathbf{A}}_{t,k}^{i} \odot \mathbf{h}_{t-1,k}^{i} + \bar{\mathbf{B}}_{t,k}^{i} \odot \mathbf{u}_{t,k}^{i},
\end{equation}
\begin{equation}
    \mathbf{y}_{t,k}^{i} = \mathbf{R}_{t,k}^{i}\mathbf{h}_{t,k}^{i} + \mathbf{D}^{i}\mathbf{u}_{t,k}^{i},
\end{equation}
where $\mathbf{B}_{t,k}^{i}$, $\mathbf{R}_{t,k}^{i}$, and $\mathbf{D}^{i}$ are the remaining SSM parameters. The output context token is then obtained by
\begin{equation}
    \bar{\mathbf{c}}_{t,k}^{i} = \mathbf{W}_o (\mathbf{y}_{t,k}^{i} \odot \phi(\mathbf{v}_{t,k}^{i})),
    \label{eq:sacp_out}
\end{equation}
where $\mathbf{v}_{t,k}^{i}$ is the corresponding token in $\mathbf{V}_t^{i}$, $\phi(\cdot)$ denotes the SiLU activation function, and $\mathbf{W}_o$ is a learnable output projection. Stacking all output tokens yields $\bar{\mathbf{C}}_t^{i}$. Because semantics are injected by modulating the scan dynamics rather than by introducing heavy cross-modal interaction, SACP preserves the efficiency of SSM-based tracking while improving robustness in cluttered anti-UAV scenarios.

\subsection{Loss Function}
\label{sec:loss}
For training, we optimize the network with a localization loss and a semantic-discriminative contrastive loss. The localization loss is defined as
\begin{equation}
    \mathcal{L}_{t} = \lambda_{\mathrm{f}} \mathcal{L}_{\mathrm{focal}} + \lambda_{\mathrm{g}} \mathcal{L}_{\mathrm{GIoU}}(\hat{\mathbf{b}}_t, \mathbf{b}_t^{*}) + \lambda_{\ell_1} \mathcal{L}_{\ell_1}(\hat{\mathbf{b}}_t, \mathbf{b}_t^{*}),
\end{equation}
where $\mathcal{L}_{\mathrm{GIoU}}$ follows the generalized IoU loss~\cite{rezatofighi2019generalized}, and $\mathbf{b}_t^{*}$ denotes the ground-truth box at frame $t$.

In addition, to enhance the semantic discriminability between the target UAV and confusing background regions, we introduce SDC. Let $\mathbf{F}_{t} \in \mathbb{R}^{HW \times d}$ denote the propagated feature map. According to the ground-truth box, we first obtain a target feature $\mathbf{v}_{t}^{+}$ by masked average pooling over the target region. We then project the target feature and the semantic embedding into a shared contrastive space:
\begin{equation}
    \mathbf{z}_{t}^{v} = \psi_{v}(\mathbf{v}_{t}^{+}), \qquad \mathbf{z}_{t}^{s} = \psi_{s}(\mathbf{s}),
\end{equation}
where $\psi_{v}(\cdot)$ and $\psi_{s}(\cdot)$ are learnable projection heads for visual and semantic features, respectively. To construct hard negatives, we project all background patches into the same space and select the top-$M$ patches that are most similar to the semantic embedding:
\begin{equation}
    \mathcal{N}_{t}^{-} = \operatorname{TopK}\Big(\operatorname{sim}\big(\psi_{v}(\mathbf{v}_{t,j}^{-}), \mathbf{z}_{t}^{s}\big)\Big),
\end{equation}
where $\mathbf{v}_{t,j}^{-}$ denotes the $j$-th background patch feature and $\operatorname{sim}(\cdot,\cdot)$ denotes cosine similarity. For brevity, we define
\begin{equation}
    s_t^{+} = \operatorname{sim}(\mathbf{z}_{t}^{v}, \mathbf{z}_{t}^{s})/\tau, \qquad
    s_{t,n}^{-} = \operatorname{sim}(\mathbf{z}_{t,n}^{-}, \mathbf{z}_{t}^{s})/\tau,
\end{equation}
where $\mathbf{z}_{t,n}^{-} \in \mathcal{N}_{t}^{-}$. The contrastive loss is formulated as
\begin{equation}
    \mathcal{L}_{\mathrm{sdc}}^{t}
    = - \log
    \frac{\exp(s_t^{+})}
    {\exp(s_t^{+}) + \sum_{n=1}^{M} \exp(s_{t,n}^{-})},
\end{equation}
where $\tau$ is the temperature parameter.

The overall training objective is
\begin{equation}
    \mathcal{L} = \sum_{t=1}^{T} \left( \mathcal{L}_{t} + \lambda_{\mathrm{sdc}} \mathcal{L}_{\mathrm{sdc}}^{t} \right),
\end{equation}
where $\lambda_{\mathrm{sdc}}$ is the balancing weight and $T$ is the number of supervised search frames. By explicitly mining semantically confusing background patches, the proposed contrastive loss improves representation discriminability and further alleviates distractor interference in anti-UAV tracking.

\begin{algorithm}[t]
\caption{TADA at frame $t$}
\label{alg:tada}
\begin{algorithmic}[1]
\REQUIRE BN input $\mathbf{X}\in\mathbb{R}^{B\times C\times H\times W}$; frozen running stats $\boldsymbol{\mu}_{\mathrm{run}},\boldsymbol{\sigma}^2_{\mathrm{run}}$ and affine $\boldsymbol{\gamma},\boldsymbol{\beta}$; FIFO queues $\mathcal{Q}_\mu,\mathcal{Q}_\sigma$ of size $L$; top ratio $\rho$; base blend $\lambda_0=n/(n+N)$; output blend $\alpha$
\ENSURE adapted output $\mathbf{Y}$
\STATE $e_{h,w}\leftarrow \frac{1}{C}\sum_{c}\mathbf{X}_{c,h,w}^2$ \COMMENT{per-location energy}
\STATE $\Omega_t\leftarrow$ top-$\rho$ locations with the highest $e_{h,w}$
\STATE $\boldsymbol{\mu}_t,\boldsymbol{\sigma}^2_t\leftarrow$ statistics computed within $\Omega_t$ only
\IF{drift-rejection test on $(\boldsymbol{\mu}_t,\boldsymbol{\sigma}^2_t)$ fails}
    \STATE reuse the previous adapted statistics \COMMENT{skip outlier frame}
\ELSE
    \STATE push $\boldsymbol{\mu}_t,\boldsymbol{\sigma}^2_t$ into $\mathcal{Q}_\mu,\mathcal{Q}_\sigma$ (pop oldest if full)
\ENDIF
\STATE $\bar{\boldsymbol{\mu}}_t\leftarrow \mathrm{mean}(\mathcal{Q}_\mu),\;\; \bar{\boldsymbol{\sigma}}^2_t\leftarrow \mathrm{mean}(\mathcal{Q}_\sigma)$ \COMMENT{temporal smoothing}
\STATE $\lambda\leftarrow \lambda_0\cdot \mathbf{c}_t$ \COMMENT{confidence gate, Eq.~(\ref{eq:tada_gate})}
\STATE $\boldsymbol{\mu}^*\leftarrow \lambda\bar{\boldsymbol{\mu}}_t+(1-\lambda)\boldsymbol{\mu}_{\mathrm{run}}$
\STATE $\boldsymbol{\sigma}^{2*}\leftarrow \lambda\bar{\boldsymbol{\sigma}}^2_t+(1-\lambda)\boldsymbol{\sigma}^2_{\mathrm{run}}$ \COMMENT{train/test interpolation}
\STATE clip $\boldsymbol{\mu}^*,\boldsymbol{\sigma}^{2*}$ to safe ranges
\STATE $\mathbf{Y}_{\mathrm{tada}}\leftarrow \boldsymbol{\gamma}\odot(\mathbf{X}-\boldsymbol{\mu}^*)/\sqrt{\boldsymbol{\sigma}^{2*}+\epsilon}+\boldsymbol{\beta}$
\STATE $\mathbf{Y}_{\mathrm{run}}\leftarrow \boldsymbol{\gamma}\odot(\mathbf{X}-\boldsymbol{\mu}_{\mathrm{run}})/\sqrt{\boldsymbol{\sigma}^2_{\mathrm{run}}+\epsilon}+\boldsymbol{\beta}$
\STATE $\mathbf{Y}\leftarrow (1-\alpha)\,\mathbf{Y}_{\mathrm{run}}+\alpha\,\mathbf{Y}_{\mathrm{tada}}$ \COMMENT{output blending, Eq.~(\ref{eq:tada_outblend})}
\RETURN $\mathbf{Y}$ \COMMENT{applied to the center and size BN branches}
\end{algorithmic}
\end{algorithm}

\subsection{Temporal-Aware Distribution Alignment}
\label{sec:tada}

Tracking models trained on large-scale generic datasets often suffer from feature distribution shifts at test time in anti-UAV scenarios with distinct illumination, background clutter, and atmospheric conditions~\cite{li2024positive,zhang2025variational,zaveri2025improving,shao2025pura}. To bridge this gap without retraining, we propose TADA, a lightweight test-time adaptation strategy that recalibrates Batch Normalization (BN) statistics on the fly while leaving all learned parameters frozen. The overall procedure is summarized in Algorithm~\ref{alg:tada}.

Given an input feature map $\mathbf{X} \in \mathbb{R}^{B \times C \times H \times W}$, standard BN normalizes with the training-time running statistics $\boldsymbol{\mu}_{\mathrm{run}}, \boldsymbol{\sigma}^2_{\mathrm{run}} \in \mathbb{R}^{C}$:
\begin{equation}
\hat{\mathbf{X}} = \frac{\mathbf{X} - \boldsymbol{\mu}_{\mathrm{run}}}{\sqrt{\boldsymbol{\sigma}^2_{\mathrm{run}} + \epsilon}},
\end{equation}
which fails to generalize once the test distribution shifts. A naive substitution by per-frame statistics computed over the entire feature map is unreliable, however, because the small UAV occupies only a few locations and the statistics are easily dominated by background clutter. We therefore estimate the per-frame statistics from the most informative locations only. For each spatial location of $\mathbf{X}$ we measure its energy $e_{h,w} = \frac{1}{C}\sum_{c} \mathbf{X}_{c,h,w}^2$, keep the top-$\rho$ fraction of locations with the highest energy as a set $\Omega_t$, and compute the current-frame statistics $\boldsymbol{\mu}_t, \boldsymbol{\sigma}^2_t$ only within $\Omega_t$. Even so, single-frame estimates remain noisy under rapid illumination changes and transient occlusions, so we maintain two FIFO queues of size $L$ over the recent per-frame statistics,
\begin{equation}
\mathcal{Q}_{\mu} = \{\boldsymbol{\mu}_{t-L+1}, \ldots, \boldsymbol{\mu}_t\}, \quad
\mathcal{Q}_{\sigma} = \{\boldsymbol{\sigma}^2_{t-L+1}, \ldots, \boldsymbol{\sigma}^2_t\},
\label{eq:tada_queue}
\end{equation}
and obtain temporally smoothed estimates by simple averaging,
\begin{equation}
\bar{\boldsymbol{\mu}}_t = \frac{1}{|\mathcal{Q}_{\mu}|}\sum_{\boldsymbol{\mu} \in \mathcal{Q}_{\mu}} \boldsymbol{\mu}, \quad
\bar{\boldsymbol{\sigma}}^2_t = \frac{1}{|\mathcal{Q}_{\sigma}|}\sum_{\boldsymbol{\sigma}^2 \in \mathcal{Q}_{\sigma}} \boldsymbol{\sigma}^2,
\label{eq:tada_smooth}
\end{equation}
which suppress transient noise while tracking the evolving target distribution. The adapted statistics are then a convex combination of the smoothed test estimates and the training statistics,
\begin{equation}
\boldsymbol{\mu}^* = \lambda \bar{\boldsymbol{\mu}}_t + (1 - \lambda) \boldsymbol{\mu}_{\mathrm{run}}, \quad
\boldsymbol{\sigma}^{2*} = \lambda \bar{\boldsymbol{\sigma}}^2_t + (1 - \lambda) \boldsymbol{\sigma}^2_{\mathrm{run}},
\label{eq:tada_interp}
\end{equation}
where the base mixing weight $\lambda_0 = n / (n + N)$ controls the trade-off: a smaller value favors training stability, while a larger one pushes the normalization toward the test-time distribution. To avoid over-adapting on ambiguous frames, we gate this weight by a per-channel confidence factor $\mathbf{c}_t \in (0,1]$ that decays as the smoothed mean departs from the training mean,
\begin{equation}
\mathbf{c}_t = \exp\!\Big(-\big(\,|\bar{\boldsymbol{\mu}}_t - \boldsymbol{\mu}_{\mathrm{run}}| / (\kappa\,\boldsymbol{\sigma}_{\mathrm{run}} + \epsilon)\,\big)^2\Big), \qquad \lambda = \lambda_0\,\mathbf{c}_t,
\label{eq:tada_gate}
\end{equation}
where $\kappa$ is a scale hyperparameter and $\boldsymbol{\sigma}_{\mathrm{run}} = \sqrt{\boldsymbol{\sigma}^2_{\mathrm{run}}}$; the gated $\lambda$ is the weight used in Eq.~(\ref{eq:tada_interp}). The adapted output applies the recalibrated normalization followed by the inherited affine transform,
\begin{equation}
\mathbf{Y}_{\mathrm{tada}} = \boldsymbol{\gamma} \odot \frac{\mathbf{X} - \boldsymbol{\mu}^*}{\sqrt{\boldsymbol{\sigma}^{2*} + \epsilon}} + \boldsymbol{\beta},
\end{equation}
where $\boldsymbol{\gamma}, \boldsymbol{\beta} \in \mathbb{R}^{C}$ are the learnable affine parameters inherited from training. Rather than overwriting the original normalization, we return a convex blend of the adapted output and the standard BN output $\mathbf{Y}_{\mathrm{run}} = \boldsymbol{\gamma} \odot \hat{\mathbf{X}} + \boldsymbol{\beta}$ obtained from the frozen running statistics,
\begin{equation}
\mathbf{Y} = (1 - \alpha)\,\mathbf{Y}_{\mathrm{run}} + \alpha\,\mathbf{Y}_{\mathrm{tada}},
\label{eq:tada_outblend}
\end{equation}
where the output coefficient $\alpha \in [0,1]$ balances stability and adaptation: $\alpha = 0$ recovers standard BN, while $\alpha = 1$ uses the fully adapted output. We apply this adaptation to the BN layers of both the center and size branches of the localization head, since test-time distribution shifts affect target localization and scale estimation jointly. We additionally adopt two safeguards detailed in the supplementary material: a drift-rejection test that skips clearly outlying frames before they enter the queues, and clipping on the adapted statistics and outputs. TADA is feed-forward, gradient-free, and incurs negligible cost, yet aligns the feature distribution with the current anti-UAV video through temporal smoothing, train/test interpolation, and output blending, while preserving model stability.

\begin{table*}[!t]
\centering
\caption{Overall comparison with state-of-the-art trackers on three anti-UAV tracking benchmarks. AUC denotes success AUC, OP50 and OP75 denote overlap precision at thresholds 0.50 and 0.75, Pre denotes precision, and nPre denotes normalized precision. Avg. AUC and Avg. Pre are computed over all three benchmarks when the corresponding scores are available. All results are reported in percentages. The $\Delta$ row reports the gap between Ours and the best competing tracker in each column, shown in \textcolor{rankfirst}{red} when Ours is higher and \textcolor{ranksecond}{blue} when Ours is lower.}
\label{tab:main_results}
\setlength{\tabcolsep}{1.4pt}
\renewcommand{\arraystretch}{1.10}
\scriptsize
\begin{tabular*}{0.98\textwidth}{@{\extracolsep{\fill}}llccccccccccc@{}}
\toprule
\multirow{2}{*}{Tracker} & \multirow{2}{*}{Venue} & \multicolumn{5}{c}{UAV-Anti-UAV~\cite{zhang2025far}} & \multicolumn{2}{c}{Anti-UAV318~\cite{jiang2021anti}} & \multicolumn{2}{c}{DUT Anti-UAV~\cite{zhao2022vision}} & \multicolumn{2}{c}{Average} \\
\cmidrule(lr){3-7} \cmidrule(lr){8-9} \cmidrule(lr){10-11} \cmidrule(lr){12-13}
 & & AUC(\%)$\uparrow$ & OP50(\%)$\uparrow$ & OP75(\%)$\uparrow$ & Pre(\%)$\uparrow$ & nPre(\%)$\uparrow$ & AUC(\%)$\uparrow$ & Pre(\%)$\uparrow$ & AUC(\%)$\uparrow$ & Pre(\%)$\uparrow$ & AUC(\%)$\uparrow$ & Pre(\%)$\uparrow$ \\
\midrule
\multicolumn{13}{l}{\emph{Pure vision trackers}} \\
SiamMask~\cite{wang2019fast} & CVPR'19 & 18.50 & 22.02 & 8.17 & 29.48 & 36.04 & 38.60 & 49.90 & 44.70 & 64.00 & 33.93 & 47.79 \\
SiamRPN++~\cite{li2019siamrpn++} & CVPR'19 & 18.81 & 22.46 & 9.94 & 29.16 & 36.59 & 45.70 & 57.60 & 58.20 & 77.00 & 40.90 & 54.59 \\
SiamFC++~\cite{XuWLYY20} & AAAI'20 & 20.35 & 23.91 & 13.18 & 29.40 & 38.40 & 44.00 & 56.10 & 52.50 & 72.40 & 38.95 & 52.63 \\
SiamBAN~\cite{ChenZLZJ20} & CVPR'20 & 19.53 & 22.53 & 12.55 & 29.37 & 37.15 & 40.50 & 52.70 & 51.40 & 69.30 & 37.14 & 50.46 \\
SiamCAR~\cite{guo2020siamcar} & CVPR'20 & 20.47 & 23.84 & 11.07 & 31.71 & 38.02 & 48.10 & 61.40 & 52.60 & 70.30 & 40.39 & 54.47 \\
LightTrack~\cite{yan2021lighttrack} & CVPR'21 & 19.31 & 20.78 & 10.58 & 27.80 & 35.66 & 44.90 & 55.90 & 47.50 & 60.60 & 37.24 & 48.10 \\
SiamGAT~\cite{guo2021graph} & CVPR'21 & 17.14 & 19.31 & 7.37 & 27.33 & 30.89 & 42.90 & 58.30 & 52.80 & 72.90 & 37.61 & 52.84 \\
AutoMatch~\cite{zhang2021learn} & ICCV'21 & 21.14 & 24.63 & 13.01 & 31.39 & 38.88 & 47.50 & 59.70 & 57.80 & 74.10 & 42.15 & 55.06 \\
HiFT~\cite{cao2021hift} & ICCV'21 & 17.09 & 19.30 & 8.26 & 25.50 & 33.91 & 41.70 & 54.20 & 39.70 & 55.30 & 32.83 & 45.00 \\
TCTrack~\cite{cao2022tctrack} & CVPR'22 & 17.24 & 19.98 & 9.10 & 27.06 & 33.33 & 42.40 & 55.20 & 46.10 & 64.10 & 35.25 & 48.79 \\
OSTrack~\cite{ye2022joint} & ECCV'22 & 29.63 & 34.38 & 24.87 & 39.37 & 47.54 & \underline{67.22} & \underline{85.40} & 61.68 & 82.75 & 52.84 & 69.17 \\
Aba-ViTrack~\cite{li2023adaptive} & ICCV'23 & 28.10 & 32.46 & 22.29 & 37.40 & 45.67 & 53.50 & 66.50 & 60.10 & 80.00 & 47.23 & 61.30 \\
ODTrack~\cite{zheng2024odtrack} & AAAI'24 & 36.17 & 42.58 & 30.38 & 48.37 & 55.43 & 63.69 & 83.36 & 64.68 & \textbf{87.39} & 54.85 & 73.04 \\
MambaNUT~\cite{wu2024mambanut} & IROS'25 & 25.12 & 29.50 & 18.59 & 34.80 & 43.09 & 53.90 & 66.80 & 54.00 & 69.90 & 44.34 & 57.17 \\
SGLATrack~\cite{xue2025similarity} & CVPR'25 & 25.49 & 29.06 & 20.22 & 33.94 & 42.41 & 54.72 & 67.83 & 55.76 & 74.78 & 45.32 & 58.85 \\
ORTrack~\cite{wu2025learning} & CVPR'25 & 26.71 & 30.70 & 20.84 & 36.70 & 44.32 & 58.80 & 74.10 & 57.60 & 77.60 & 47.70 & 62.80 \\
MambaLCT~\cite{li2025mambalct} & AAAI'25 & 38.88 & 45.29 & \underline{33.03} & 51.97 & 57.97 & 64.24 & \textbf{85.87} & 62.95 & 84.98 & 55.36 & 74.27 \\
MCITrack-B224~\cite{kang2025exploring} & AAAI'25 & 34.90 & 41.23 & 27.92 & 47.54 & 54.17 & 66.51 & 82.76 & \underline{65.93} & 84.37 & 55.78 & 71.56 \\
\midrule
\multicolumn{13}{l}{\emph{Vision-language trackers}} \\
CiteTracker-256~\cite{li2023citetracker} & ICCV'23 & 26.78 & 30.74 & 21.76 & 35.64 & 44.10 & 54.50 & 67.00 & 55.30 & 71.90 & 45.53 & 58.18 \\
UVLTrack~\cite{ma2024unifying} & AAAI'24 & 33.78 & 40.41 & 26.11 & 46.89 & 53.91 & 63.95 & 80.62 & 60.31 & 80.73 & 52.68 & 69.41 \\
MambaTrack~\cite{zhang2025mambatrack} & ICASSP'25 & 27.69 & 32.29 & 21.28 & 38.07 & 45.53 & 57.67 & 73.00 & 61.99 & 84.91 & 49.12 & 65.33 \\
DUTrack-256~\cite{li2025dynamic} & CVPR'25 & 38.61 & 45.88 & 31.77 & 52.37 & 59.15 & 56.93 & 74.72 & 63.84 & 86.66 & 53.13 & 71.25 \\
ATCTrack~\cite{feng2025atctrack} & ICCV'25 & 37.24 & 44.96 & 29.56 & 50.63 & 58.42 & 65.03 & 80.54 & 63.59 & 81.53 & 55.29 & 70.90 \\
MambaSTS~\cite{zhang2025far} & arXiv'25 & \underline{44.03} & \underline{54.36} & \underline{33.03} & \underline{61.35} & \underline{68.30} & 62.50 & 84.40 & 61.30 & 82.20 & \underline{55.94} & \underline{75.98} \\
\textbf{Ours} & - & \textbf{46.40} & \textbf{54.94} & \textbf{38.63} & \textbf{62.39} & \textbf{68.64} & \textbf{67.27} & 84.13 & \textbf{66.36} & \underline{87.11} & \textbf{60.01} & \textbf{77.88} \\
\midrule
$\Delta$ & & \textcolor{rankfirst}{\textbf{+2.37}} & \textcolor{rankfirst}{\textbf{+0.58}} & \textcolor{rankfirst}{\textbf{+5.60}} & \textcolor{rankfirst}{\textbf{+1.04}} & \textcolor{rankfirst}{\textbf{+0.34}} & \textcolor{rankfirst}{\textbf{+0.05}} & \textcolor{ranksecond}{\textbf{-1.74}} & \textcolor{rankfirst}{\textbf{+0.43}} & \textcolor{ranksecond}{\textbf{-0.28}} & \textcolor{rankfirst}{\textbf{+4.07}} & \textcolor{rankfirst}{\textbf{+1.90}} \\
\bottomrule
\end{tabular*}
\end{table*}

%% file: section/4_exp.tex
In this section, we describe the datasets, evaluation metrics, and implementation details, and then present comparisons with state-of-the-art trackers, ablation studies, and qualitative analyses.

\begin{table}[t]
\centering
\caption{Attribute-based AUC comparison on UAV-Anti-UAV. The $\Delta$ row reports the gap between Ours and the best competing tracker in each column, shown in \textcolor{rankfirst}{red} when Ours is higher and \textcolor{ranksecond}{blue} when Ours is lower.}
\label{tab:attribute_results}
\setlength{\tabcolsep}{2.4pt}
\renewcommand{\arraystretch}{1.1}
\footnotesize
\begin{tabular}{lccccccc}
\toprule
Tracker & FM & MB & SO & OV & SD & ARV & Avg. \\
\midrule
MambaLCT~\cite{li2025mambalct} & 33.12 & 32.11 & 24.69 & 29.27 & 32.33 & 36.98 & 31.42 \\
MCITrack-B224~\cite{kang2025exploring} & 29.71 & 28.68 & 21.48 & 26.94 & 28.81 & 33.37 & 28.17 \\
DUTrack-256~\cite{li2025dynamic} & 32.68 & 31.43 & 23.65 & 29.00 & 32.71 & 37.25 & 31.12 \\
MambaSTS~\cite{zhang2025far} & \underline{38.55} & \underline{37.08} & \underline{30.66} & \underline{34.31} & \underline{38.28} & \underline{42.82} & \underline{36.95} \\
\textbf{Ours} & \textbf{40.19} & \textbf{39.88} & \textbf{31.67} & \textbf{35.06} & \textbf{41.32} & \textbf{45.10} & \textbf{38.87} \\
\midrule
$\Delta$ & \textcolor{rankfirst}{\textbf{+1.64}} & \textcolor{rankfirst}{\textbf{+2.80}} & \textcolor{rankfirst}{\textbf{+1.01}} & \textcolor{rankfirst}{\textbf{+0.75}} & \textcolor{rankfirst}{\textbf{+3.04}} & \textcolor{rankfirst}{\textbf{+2.28}} & \textcolor{rankfirst}{\textbf{+1.92}} \\
\bottomrule
\end{tabular}
\end{table}

\subsection{Datasets and Evaluation Metrics}
\label{sec:datasets}

\noindent\textbf{Datasets.}
We evaluate the proposed framework on three anti-UAV tracking benchmarks: Anti-UAV318~\cite{jiang2021anti}, DUT Anti-UAV~\cite{zhao2022vision}, and UAV-Anti-UAV~\cite{zhang2025far}. Anti-UAV318 is a large-scale RGB-T benchmark for ground-to-air UAV tracking with visible and invisible target states. DUT Anti-UAV provides visible-light detection and tracking subsets with precise annotations for small UAV targets in short- and long-term videos. UAV-Anti-UAV targets the UAV-to-UAV tracking setting, where both the camera platform and target UAV undergo rapid motion, resulting in stronger dual-dynamic disturbance, background change, and target scale variation. We adopt a two-stage training protocol: the tracker is first pretrained on the combination of LaSOT~\cite{fan2019lasot}, GOT-10k~\cite{huang2019got}, TrackingNet~\cite{muller2018trackingnet}, and COCO~\cite{lin2014microsoft} with equal sampling ratios to acquire general appearance and matching representations, and then finetuned on the anti-UAV training data together with the natural-language target description.

\begin{table}[t]
\centering
\caption{Generalization comparison on UAV tracking benchmarks. The $\Delta$ row reports the gap between Ours and the best competing tracker in each column, shown in \textcolor{rankfirst}{red} when Ours is higher and \textcolor{ranksecond}{blue} when Ours is lower.}
\label{tab:uav_generalization}
\setlength{\tabcolsep}{0pt}
\renewcommand{\arraystretch}{1.1}
\footnotesize
\begin{tabular*}{0.90\columnwidth}{@{\extracolsep{\fill}}lcccc@{}}
\toprule
\multirow{2}{*}{Tracker} & \multicolumn{2}{c}{UAV123~\cite{mueller2016benchmark}} & \multicolumn{2}{c}{UAVDT~\cite{du2018unmanned}} \\
\cmidrule(lr){2-3} \cmidrule(lr){4-5}
 & AUC(\%)$\uparrow$ & Pre(\%)$\uparrow$ & AUC(\%)$\uparrow$ & Pre(\%)$\uparrow$ \\
\midrule
SGLATrack~\cite{xue2025similarity} & 66.73 & 87.22 & 58.96 & 80.66 \\
MambaLCT~\cite{li2025mambalct} & 69.15 & \underline{90.97} & \underline{64.93} & 85.14 \\
MCITrack~\cite{kang2025exploring} & \underline{69.75} & 90.48 & 63.23 & 82.76 \\
DUTrack~\cite{li2025dynamic} & 68.10 & 90.77 & 64.83 & \textbf{87.73} \\
MambaTrack~\cite{zhang2025mambatrack} & 66.82 & 89.60 & 57.05 & 80.30 \\
\textbf{Ours} & \textbf{70.28} & \textbf{92.31} & \textbf{65.80} & \underline{86.93} \\
\midrule
$\Delta$ & \textcolor{rankfirst}{\textbf{+0.53}} & \textcolor{rankfirst}{\textbf{+1.34}} & \textcolor{rankfirst}{\textbf{+0.87}} & \textcolor{ranksecond}{\textbf{-0.80}} \\
\bottomrule
\end{tabular*}
\end{table}

\noindent\textbf{Evaluation Metrics.}
Following the one-pass evaluation (OPE) protocol, we report AUC and precision (Pre) on Anti-UAV318 and DUT Anti-UAV. For UAV-Anti-UAV, we follow its official protocol and report five metrics: success AUC, overlap precision at thresholds 0.50 and 0.75 (OP50 and OP75), precision (Pre), and normalized precision (nPre). All metrics are reported in percentages, and higher values indicate better performance.

\subsection{Implementation Details}

\noindent\textbf{Model Configuration.}
We adopt Fast-iTPN-Base~\cite{tian2024fast} as the visual backbone, initialized from its CLIP-distilled pretraining, with a patch stride of $16$ and a drop-path rate of $0.1$. The template and search region are resized to $112\!\times\!112$ and $224\!\times\!224$ with context factors of $2.0$ and $4.0$, respectively, and three templates plus two search frames are sampled per training pair within a temporal window of $400$ frames. The SACP module is inserted at four interaction points across the backbone with a $512$-dimensional neck, and a frozen CLIP~\cite{radford2021learning} ViT-B/32 text encoder produces the $512$-dim semantic embedding that modulates the visual stream in a gated manner.

\noindent\textbf{Training.}
The framework is optimized in two stages with AdamW (weight decay $10^{-4}$, gradient clipping $0.1$) and a step learning-rate schedule. In Stage~I, the tracker is pretrained for $300$ epochs with a base learning rate of $4\!\times\!10^{-4}$ and a backbone multiplier of $0.1$, decayed by $10$ at epoch $240$. In Stage~II, the model is finetuned on the UAV-Anti-UAV training split for $100$ epochs with the base learning rate reduced to $1\!\times\!10^{-4}$ and dropped at epoch $80$; all BN layers are frozen so that the Stage-I running statistics are preserved for test-time adaptation by TADA, and the semantic-discriminative contrastive loss is enabled with a weight of $0.2$, while the focal, GIoU, and $\ell_1$ losses use weights of $1.0$, $2.0$, and $5.0$, respectively. All training is performed on four NVIDIA H100 GPUs with a total batch size of $64$.

\begin{figure*}[!t]
    \centering
    \includegraphics[width=\textwidth]{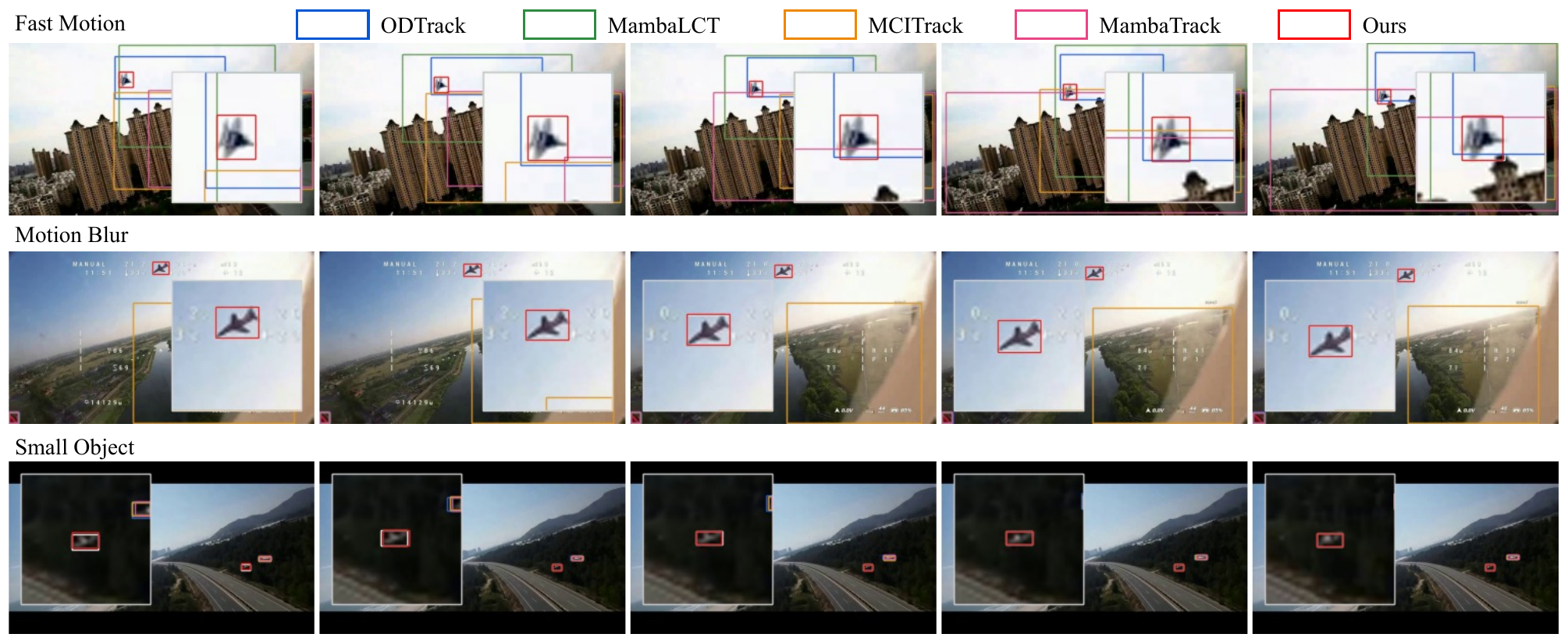}
    \caption{Qualitative comparison with ODTrack, MambaLCT, MCITrack, and MambaTrack on UAV-Anti-UAV sequences involving fast motion, motion blur, and small objects. Our method maintains tighter and more stable localization when baseline trackers drift, lag behind, or produce loose boxes.}
    \label{fig:qualitative_results}
\end{figure*}

\begin{figure*}[!t]
    \centering
    \includegraphics[width=\textwidth]{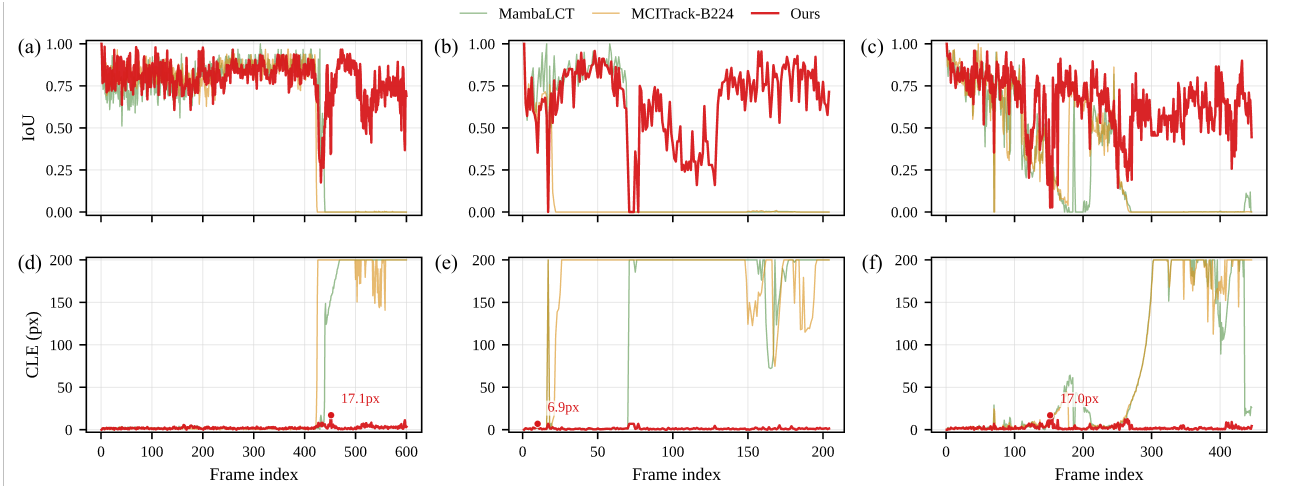}
    \caption{Temporal stability comparison on three representative UAV-Anti-UAV sequences. The first row shows frame-wise IoU curves, and the second row shows center location error (CLE) curves. Our tracker maintains smoother IoU responses and lower CLE values than MambaLCT and MCITrack-B224.}
    \label{fig:temporal_stability}
\end{figure*}

\begin{table}[t]
\centering
\caption{Efficiency comparison on the UAV-Anti-UAV benchmark.}
\label{tab:efficiency}
\setlength{\tabcolsep}{2.0pt}
\renewcommand{\arraystretch}{1.1}
\footnotesize
\begin{tabular}{llccc}
\toprule
Tracker & Backbone & AUC(\%)$\uparrow$ & Params(M) & FPS$\uparrow$ \\
\midrule
ODTrack~\cite{zheng2024odtrack} & ViT-B & 36.17 & \underline{92.83} & 38.61 \\
MambaLCT~\cite{li2025mambalct} & HiViT-B & \underline{38.88} & \textbf{72.78} & \textbf{97.08} \\
MCITrack-B224~\cite{kang2025exploring} & Fast-iTPN-B & 34.90 & 107.15 & 43.32 \\
MambaTrack~\cite{zhang2025mambatrack} & Vim-S & 27.69 & 221.47 & \underline{48.11} \\
\textbf{Ours} & Fast-iTPN-B & \textbf{46.40} & 189.87 & 41.91 \\
\bottomrule
\end{tabular}
\end{table}

\subsection{Comparison with State-of-the-Art Methods}

\subsubsection{Overall Performance}
We compare the proposed tracker with representative pure-vision and vision-language trackers in Tab.~\ref{tab:main_results}. On UAV-Anti-UAV, our method ranks first on all five metrics, achieving $46.40\%$ AUC and $38.63\%$ OP75. Compared with the strongest competitor MambaSTS, it improves AUC by $+2.37\%$ and, most notably, OP75 by $+5.60\%$, suggesting more accurate box localization under fast relative motion and frequent scale changes. The advantage over MambaLCT and MCITrack indicates that SACP is more robust than purely visual temporal state modeling in UAV-to-UAV tracking. On Anti-UAV318 and DUT Anti-UAV, our method also obtains the best AUC scores of $67.27\%$ and $66.36\%$, respectively. ODTrack attains a slightly higher precision on DUT Anti-UAV, whose ground-based sequences contain larger and slower targets that favor its video-level temporal token propagation; our gains are instead concentrated on the harder UAV-to-UAV setting, and our method still achieves the best average performance over the three benchmarks, with $60.01\%$ AUC and $77.88\%$ precision. This cross-benchmark consistency shows that the proposed framework benefits both UAV-mounted and ground-based anti-UAV tracking, where domain shift, small targets, and distractors remain common challenges.

\subsubsection{Attribute-Based Analysis}

Attribute-level results in Tab.~\ref{tab:attribute_results} provide a finer view of where the proposed tracker improves over existing UAV-to-UAV trackers. Our method obtains the best average AUC of $38.87\%$, surpassing the strongest competitor MambaSTS by $+1.92\%$, and ranks first on all six attributes: fast motion (FM), motion blur (MB), small object (SO), out-of-view (OV), similar distractor (SD), and aspect-ratio variation (ARV). This all-attribute lead is important because UAV-Anti-UAV sequences rarely contain an isolated challenge; rapid ego-motion often co-occurs with blur, scale change, and background distractors. Therefore, the result suggests that the proposed semantic-aware temporal propagation improves the overall stability of the tracker rather than only overfitting to a single failure mode. The largest gain appears on similar distractors, where our method reaches $41.32\%$ AUC and improves over MambaSTS by $+3.04\%$. This is the most direct empirical signature of the SACP mechanism in Sec.~\ref{sec:context}: visually similar UAV-like regions tend to be written into the temporal memory of pure-vision trackers and gradually contaminate the propagated state, whereas the semantic write gate in SACP down-weights such distractor tokens at the moment of state update, so the propagated context stays anchored to the described target rather than to look-alike background. That the attribute most directly tied to target-identity confusion benefits the most is consistent with our central claim that semantics should control \emph{what} the temporal memory retains, not merely match appearance within a single frame. The gains on motion blur ($39.88\%$, $+2.80\%$) and aspect-ratio variation ($45.10\%$, $+2.28\%$) further show that the tracker remains reliable when frame-level appearance is degraded or box geometry changes sharply. In these cases, propagated semantic context helps preserve identity, while TADA reduces video-specific feature-statistic mismatch caused by changing viewpoints and background distributions. For fast motion, the improvement of $+1.64\%$ indicates that temporal context alleviates the lagging and response-shift problem caused by large inter-frame displacement. For out-of-view, the gain is smaller ($+0.75\%$), which is reasonable because no tracker can fully recover missing visual evidence without an explicit long-term re-detection branch; nevertheless, the improvement suggests that the propagated state remains useful when the target is only partially absent or quickly reappears. The smallest gain is observed on small objects ($+1.01\%$), where the UAV occupies only a few pixels and both semantic and visual cues have limited spatial evidence to exploit. This remaining gap reveals a limitation of the current framework and motivates future designs with stronger tiny-target localization or re-detection mechanisms.

\begin{figure*}[!t]
    \centering
    \includegraphics[width=0.98\textwidth]{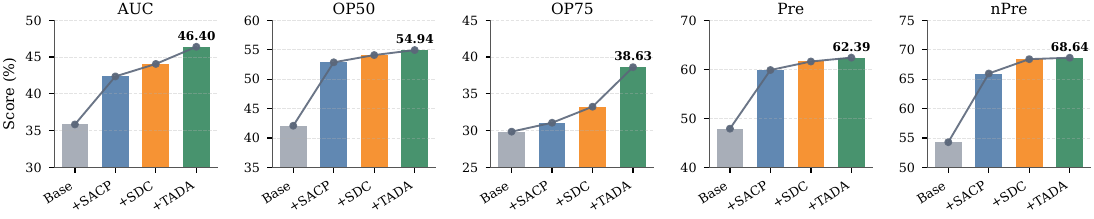}
    \caption{Bar-chart visualization of the component ablation on UAV-Anti-UAV. Each panel shows one metric under progressively added components, from the baseline to the full model.}
    \label{fig:component_ablation_bar}
\end{figure*}

\subsubsection{Qualitative Results}

We provide qualitative comparisons with ODTrack, MambaLCT, MCITrack, and MambaTrack on representative UAV-to-UAV scenarios in Fig.~\ref{fig:qualitative_results}. The selected cases cover fast motion, motion blur, and small objects, which are common failure modes in UAV-Anti-UAV tracking and also correspond to the attributes analyzed in Tab.~\ref{tab:attribute_results}. Under fast motion, several baselines lag behind the target or shift toward nearby high-response background regions because the inter-frame displacement is large and the search region changes rapidly. Under motion blur, the target boundary becomes less discriminative, causing loose boxes or box-center deviation. For small objects, the target occupies only a few pixels, so local texture and shape cues are weak and the tracker can easily confuse the UAV with clutter or structured background patterns. In contrast, our tracker produces boxes that remain more tightly aligned with the target across these representative cases. The improvement is visible not only in the box center but also in box scale: the predicted boxes are less likely to expand onto background regions or collapse around partial target responses. This behavior is consistent with the design of SACP, where the target description is used to guide temporal context propagation and preserve identity when visual evidence becomes unstable. TADA further helps keep the localization head calibrated to the current video, which is useful when blur, illumination, or viewpoint changes alter the feature distribution. Therefore, the qualitative results provide a visual explanation for the attribute-level gains: the proposed method does not merely produce occasional better boxes, but maintains target-consistent localization under the typical degradation patterns of UAV-to-UAV tracking.

\begin{figure*}[!t]
    \centering
    \includegraphics[width=\textwidth]{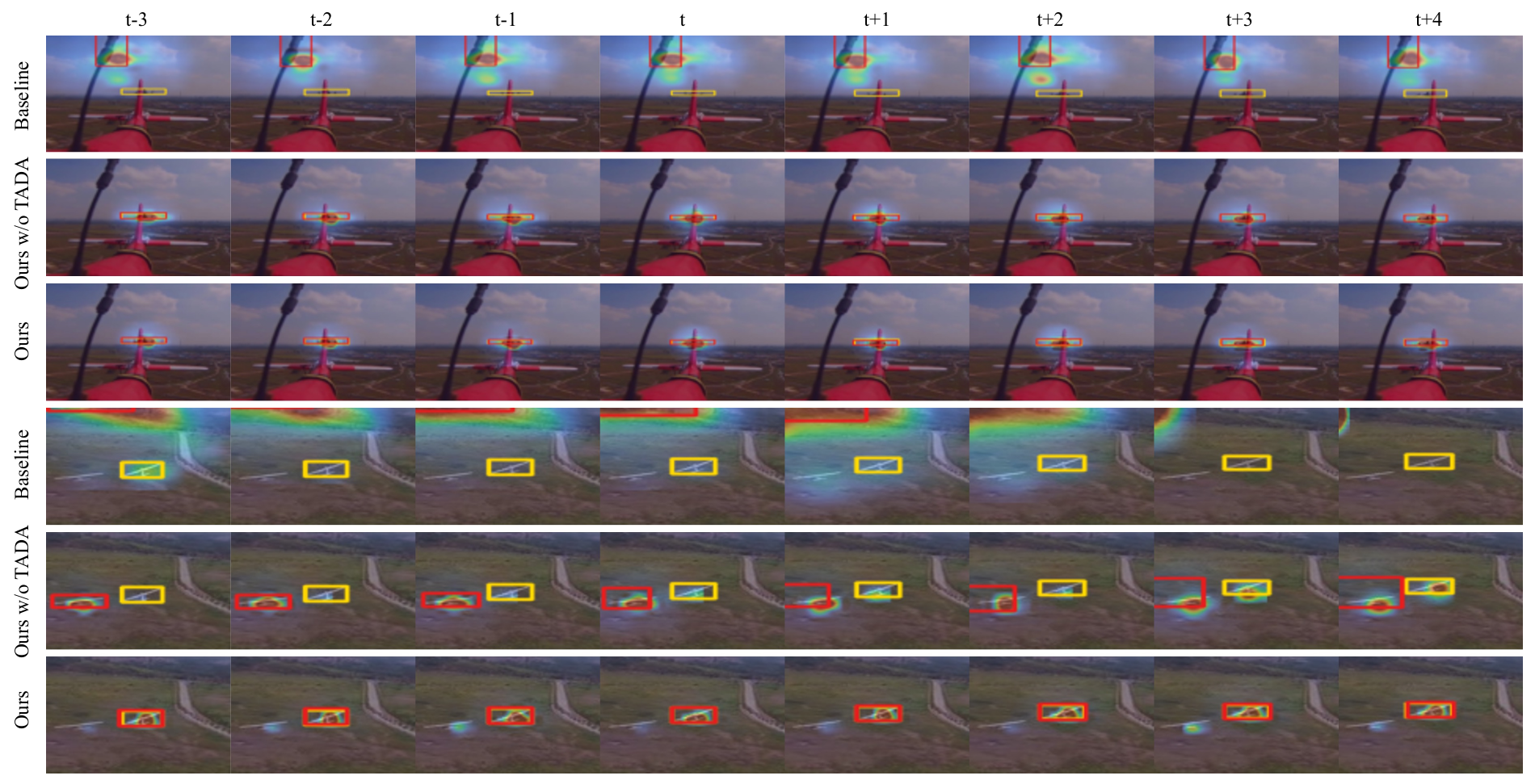}
    \caption{Response-map visualization over short UAV-Anti-UAV video clips (eight consecutive frames each), comparing the pure-vision baseline, Ours w/o TADA, and the full model. The baseline progressively leaks response onto distractor regions, while SACP keeps it anchored to the target. Yellow boxes denote ground truth, and red boxes denote predictions.}
    \label{fig:response_map}
\end{figure*}

\subsubsection{Temporal Stability Analysis}

To evaluate temporal robustness, we visualize the frame-wise IoU and center location error (CLE) on three representative UAV-Anti-UAV sequences in Fig.~\ref{fig:temporal_stability}. IoU reflects the quality of box overlap, while CLE measures the center-position error; together, they reveal whether a tracker is both spatially accurate and temporally stable. Compared with MambaLCT and MCITrack-B224, our tracker maintains smoother IoU curves over long sequences and avoids abrupt collapses during challenging intervals. In contrast, the baselines frequently exhibit sudden drops in IoU accompanied by sharp increases in CLE, which indicates target drift or temporary target loss. These failures usually occur when the target undergoes rapid displacement, becomes blurred, or passes through visually confusing background regions. The curves also show that our advantage is not limited to isolated frames. Across the selected sequences, the CLE of our tracker remains below $18$ pixels, whereas the baselines often reach the clipped upper range of the plot after drift. This suggests that our method better controls error accumulation: even when individual frames are ambiguous, the propagated temporal state prevents the tracker from quickly jumping to distractors or stale background responses. SACP contributes by carrying target-aware context across frames, and TADA further stabilizes the localization head by adapting feature statistics to the current video without updating model parameters. The temporal-stability visualization therefore complements the average metrics in Tab.~\ref{tab:main_results}: the performance gain comes from fewer catastrophic drifts and more consistent localization throughout the sequence, which is critical for UAV-to-UAV tracking where downstream control decisions depend on stable frame-by-frame estimates.

\begin{figure*}[!t]
    \centering
    \includegraphics[width=\textwidth]{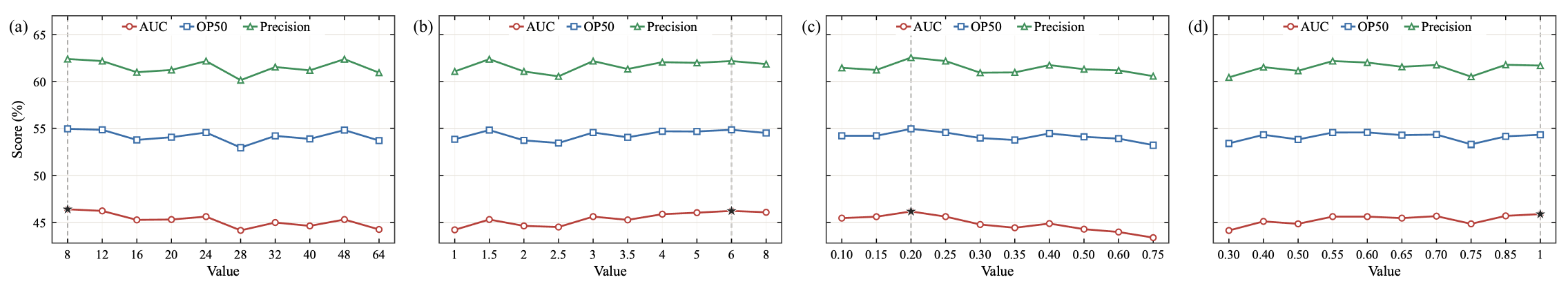}
    \caption{Hyperparameter sensitivity analysis of TADA on UAV-Anti-UAV. From left to right, (a) to (d) vary the candidate number $N$, current-frame weight $n$, output blending coefficient $\alpha$, and top-$k$ ratio $\rho$, respectively. We report AUC, OP50, and precision.}
    \label{fig:tada_hyperparameter_ablation}
\end{figure*}

\subsubsection{Generalization to UAV Tracking}

To assess whether the proposed design over-specializes to anti-UAV data, we further evaluate it on generic UAV tracking benchmarks in Tab.~\ref{tab:uav_generalization}. UAV123 contains diverse low-altitude tracking videos captured from UAV-mounted cameras, while UAVDT focuses on traffic-related aerial scenes with frequent scale changes, camera motion, and background clutter. These benchmarks differ from UAV-Anti-UAV because the tracked targets are mostly ground objects rather than flying UAVs, so strong performance on them indicates whether the learned temporal representation can transfer beyond the anti-UAV category. On UAV123, our method achieves the best AUC of $70.28\%$ and precision of $92.31\%$, outperforming MCITrack in AUC and MambaLCT in precision. On UAVDT, it obtains the best AUC of $65.80\%$ and remains close to DUTrack in precision ($86.93\%$ vs.\ $87.73\%$). The slightly lower precision on UAVDT suggests that DUTrack can still be competitive in center localization for traffic-like targets, but our higher AUC indicates better overall overlap quality and scale estimation. Overall, these results show that the proposed modules do not sacrifice generic UAV tracking ability for anti-UAV specialization. Instead, semantic-aware temporal propagation and conservative test-time alignment learn transferable robustness to UAV camera motion, scale variation, and background changes, which are shared across both anti-UAV and generic UAV tracking scenarios.

\subsubsection{Efficiency Comparison}

Besides tracking accuracy, runtime efficiency is critical for UAV-to-UAV tracking because the tracker must process rapidly changing aerial videos. As shown in Tab.~\ref{tab:efficiency}, our method achieves the highest AUC of $46.40\%$ while running at $41.91$ FPS, meeting real-time requirements. The reported $189.87$M parameters include the frozen CLIP text encoder, whereas only $126.44$M parameters are trainable; moreover, each target description is encoded once per sequence rather than per frame, so the text branch adds negligible per-frame cost. Although MambaLCT runs faster, our method improves AUC by $+7.52\%$; and compared with MambaTrack, our method uses fewer parameters ($189.87$M vs.\ $221.47$M) while obtaining substantially higher accuracy ($+18.71\%$ AUC). These results show that the proposed framework provides a favorable accuracy-efficiency trade-off for UAV-to-UAV tracking.

\subsection{Ablation Study}

We design a series of ablation studies to isolate the contribution of each component. Unless otherwise specified, all ablation results are evaluated on UAV-Anti-UAV with the same training and evaluation protocol as the main comparison.

\subsubsection{Component Analysis}

As shown in Fig.~\ref{fig:component_ablation_bar}, SACP brings the largest improvement, increasing AUC from $35.84\%$ to $42.37\%$ and precision from $47.90\%$ to $59.86\%$. This confirms that SACP is the key component for handling rapid target-camera motion. Adding SDC improves AUC to $44.07\%$ and nPre to $68.40\%$, indicating stronger target-background discrimination. With TADA, the full model reaches $46.40\%$ AUC, $62.39\%$ precision, and $68.64\%$ nPre. Overall, most of the gain comes from SACP, while SDC and TADA act as complementary refinements that sharpen discrimination and improve test-time robustness, respectively. To explain the component-level improvements, we visualize decoder response maps over short video clips in Fig.~\ref{fig:response_map}. Each clip contains eight consecutive frames, and we compare three variants: the pure-vision baseline, Ours w/o TADA, and the full model. The yellow boxes denote ground-truth annotations, while the red boxes denote predicted bounding boxes. The key pattern to read across the eight frames is temporal: the baseline gradually leaks response onto distracting regions and drifts away from the target, which is the visual footprint of a propagated state being progressively contaminated by distractor evidence. With SACP and SDC, Ours w/o TADA keeps the response anchored to the true target throughout the clip, indicating that the semantic write gate suppresses distractor tokens at the moment of state update and prevents this contamination from accumulating, rather than merely improving a single frame. Adding TADA keeps these response peaks compact and consistent across consecutive frames, especially under rapid appearance and background changes, which reflects its role in stabilizing predictions when individual frames are ambiguous. Overall, the response maps mirror the quantitative component analysis: SACP and SDC drive target discrimination, while TADA contributes cross-frame consistency at test time.

\subsubsection{SACP Design Analysis}

\begin{figure}[t]
    \centering
    \includegraphics[width=\columnwidth]{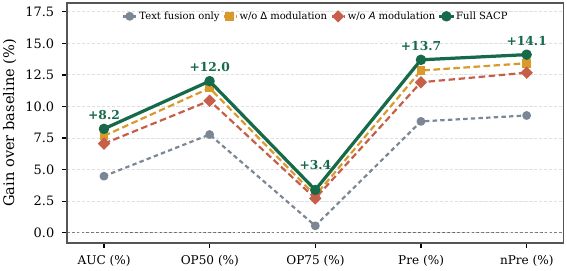}
    \caption{Ablation of SACP designs on UAV-Anti-UAV without TADA. The plot reports metric gains over the baseline for different semantic injection strategies.}
    \label{fig:sacp_design_ablation}
\end{figure}

Fig.~\ref{fig:sacp_design_ablation} compares different semantic injection designs, evaluated on the SACP+SDC configuration without TADA. The baseline reaches $35.84\%$ AUC, $42.08\%$ OP50, $29.86\%$ OP75, $47.90\%$ precision, and $54.29\%$ nPre. Direct text fusion improves these results to $40.32\%$, $49.86\%$, $30.41\%$, $56.72\%$, and $63.58\%$, respectively, showing that target semantics already provide useful discrimination cues. However, it remains clearly weaker than our full design, in which semantics modulate both the write gate $\boldsymbol{\Delta}$ and the retention matrix $\mathbf{A}$, reaching $44.07\%$ AUC, $54.09\%$ OP50, $33.25\%$ OP75, $61.60\%$ precision, and $68.40\%$ nPre. Disabling the semantic modulation of the write gate $\boldsymbol{\Delta}$ or of the state-transition matrix $\mathbf{A}$ lowers AUC to $43.51\%$ and $42.90\%$, respectively; the larger drop from removing $\mathbf{A}$ modulation indicates that controlling \emph{how long} target-consistent context is retained is at least as important as controlling \emph{what} is written into the temporal memory. Crucially, every modulation variant outperforms direct text fusion ($40.32\%$) by a clear margin, which confirms our central design principle: target semantics should govern the selection dynamics of the temporal memory—its write gate and retention timescale—rather than merely be fused as an auxiliary appearance feature.

\subsubsection{TADA Analysis}

We analyze TADA from two complementary angles: the contribution of its key components, and its sensitivity to the main hyperparameters.

\noindent\textbf{Component contribution.}
Tab.~\ref{tab:tada_component_ablation} studies the contribution of the core components in TADA. Compared with the model without TADA, the full strategy improves AUC from $44.07\%$ to $46.40\%$ and OP75 from $33.25\%$ to $38.63\%$, showing that test-time distribution alignment strengthens inference-time robustness. Among the core components, removing the top-$k$ statistics causes the largest degradation (AUC $43.34\%$, OP75 $32.59\%$): estimating the adaptation statistics only from the most informative spatial locations, rather than the entire feature map, is essential for avoiding noisy background-dominated statistics. The temporal queue in Eqs.~(\ref{eq:tada_queue}) and~(\ref{eq:tada_smooth}), which smooths per-frame statistics over a short window, mainly improves the high-overlap regime (OP75 $38.03\%$ without it). Adapting both the center and size branches of the localization head is also more effective than adapting either branch alone, as shown by the center-only and size-only variants. The remaining lightweight safeguards (confidence gate, drift rejection, statistic clipping, and output regularization) each contribute a small and consistent gain; we report their individual effects in the supplementary material.

\begin{table}[t]
\centering
\caption{Ablation study of key components in TADA on UAV-Anti-UAV.}
\label{tab:tada_component_ablation}
\setlength{\tabcolsep}{2pt}
\renewcommand{\arraystretch}{1.1}
\footnotesize
\begin{tabular}{lccccc}
\toprule
Variant & AUC(\%) & OP50(\%) & OP75(\%) & Pre(\%) & nPre(\%) \\
\midrule
w/o TADA & 44.07 & 54.09 & 33.25 & \underline{61.60} & \underline{68.40} \\
w/o top-$k$ statistics & 43.34 & 53.67 & 32.59 & 60.98 & 67.71 \\
w/o temporal queue & 45.51 & 53.82 & 38.03 & 61.01 & 67.53 \\
Center-only adaptation & 43.84 & 53.84 & 33.03 & 61.26 & 68.15 \\
Size-only adaptation & \underline{45.72} & \underline{54.14} & \underline{38.21} & 61.24 & 67.90 \\
\textbf{Full TADA} & \textbf{46.40} & \textbf{54.94} & \textbf{38.63} & \textbf{62.39} & \textbf{68.64} \\
\bottomrule
\end{tabular}
\end{table}

\noindent\textbf{Hyperparameter sensitivity.}
We further analyze the sensitivity of the proposed TADA module to four key hyperparameters on UAV-Anti-UAV, as shown in Fig.~\ref{fig:tada_hyperparameter_ablation}. For each hyperparameter, we vary its value while keeping the others fixed, and report AUC, OP50, and precision. Overall, TADA is relatively stable within a reasonable range of hyperparameter values, while overly aggressive choices may lead to performance degradation. The candidate number $N$ and current-frame weight $n$ together determine the base blending weight $\lambda_0=n/(n+N)$ in Eq.~(\ref{eq:tada_interp}). In Fig.~\ref{fig:tada_hyperparameter_ablation}(a), the candidate number $N$ achieves the best performance at $N=8$, with an AUC of $46.40\%$, OP50 of $54.94\%$, and precision of $62.39\%$. Increasing $N$ does not consistently improve the results, suggesting that introducing too many candidates may bring noisy statistics into the test-time adaptation process. In Fig.~\ref{fig:tada_hyperparameter_ablation}(b), the performance improves as the current-frame weight $n$ increases from $1$ to $6$, and then becomes saturated. This indicates that emphasizing reliable current-frame observations is beneficial, but an excessively large weight provides limited additional gain. In Fig.~\ref{fig:tada_hyperparameter_ablation}(c), larger output blending coefficients $\alpha$ in Eq.~(\ref{eq:tada_outblend}) generally lead to better AUC, and the best result in this sweep is obtained when $\alpha=1.0$. This suggests that the adapted prediction is more reliable than the original output on UAV-Anti-UAV. In Fig.~\ref{fig:tada_hyperparameter_ablation}(d), the best AUC is obtained at the top-$k$ ratio $\rho=0.20$, while larger ratios gradually reduce performance. This observation indicates that using a compact set of high-energy spatial responses is important for suppressing noisy background statistics during adaptation. Based on these observations, we use $N=8$, $n=6$, $\alpha=1.0$, and $\rho=0.20$ as the final hyperparameter values for TADA on UAV-Anti-UAV, and the final reported model follows the configuration that achieves $46.40\%$ AUC.

%% file: section/5_discussion.tex
\begin{figure}[!t]
    \centering
    \includegraphics[width=\columnwidth]{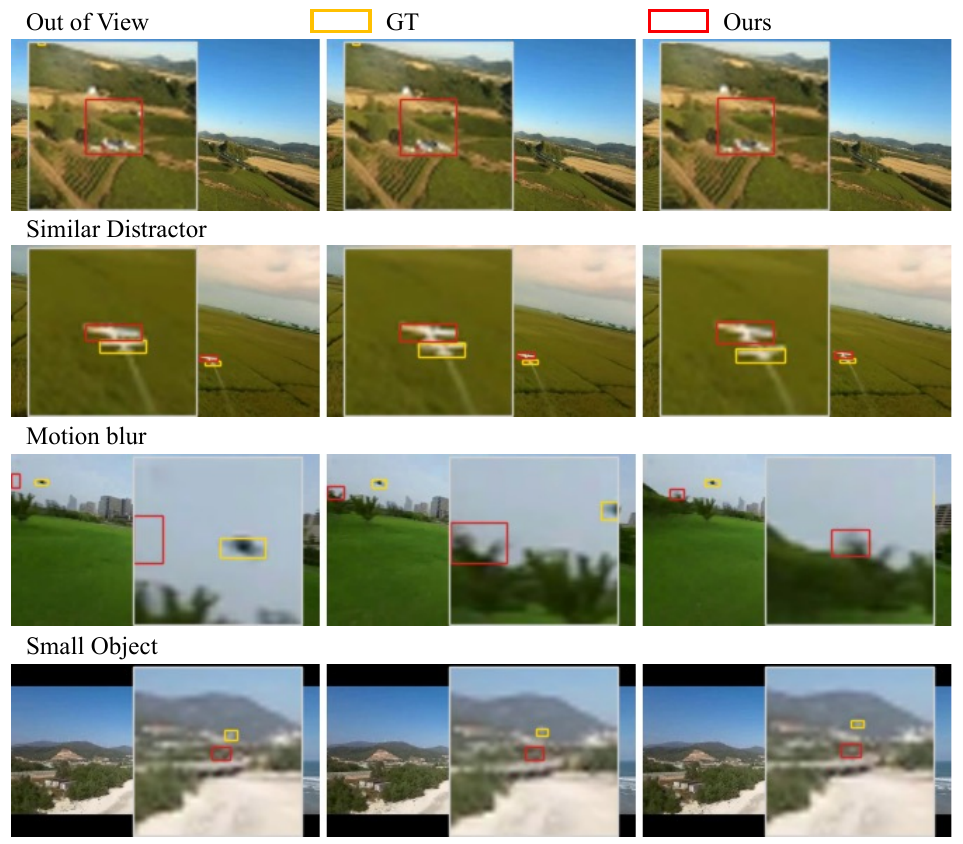}
    \caption{Representative challenging cases where our tracker may still produce inaccurate localization on UAV-Anti-UAV, including out-of-view, similar distractors, severe motion blur, and extremely small objects.}
    \label{fig:failure_cases}
\end{figure}

\subsection{Failure Cases Analysis}

Representative failure cases are illustrated in Fig.~\ref{fig:failure_cases}. Although the proposed framework improves robustness in most UAV-to-UAV tracking scenarios, failures may still occur under challenging conditions including out-of-view targets, visually similar distractors, severe motion blur, and extremely small objects. In out-of-view scenarios, the target is temporarily absent from the observation, requiring the tracker to rely on historical states and appearance priors for re-localization. When the target reappears after a long absence or from an unexpected location, such priors may become outdated, leading to inaccurate recovery in some cases. Ambiguities introduced by similar distractors remain another challenge. While semantic guidance improves discrimination against look-alike objects, multiple UAV-like instances or structured background patterns may still produce responses comparable to the true target, occasionally resulting in incorrect but visually plausible localization. Severe motion blur further degrades boundary and texture information, reducing the reliability of both classification confidence and box regression. For extremely small targets, the object occupies only a few pixels, making localization highly sensitive to small prediction errors, which can significantly affect IoU-based evaluation. These cases indicate remaining limitations rather than systematic failures. SACP is effective in preserving target identity over time but is not designed for long-term re-detection when the target is absent for extended periods. Similarly, TADA alleviates feature distribution shift but cannot recover spatial details lost under severe blur or extreme scale reduction.


\subsection{Future Work}

These limitations suggest several directions for future work. A long-term memory and re-detection branch could improve recovery after target disappearance, while instance-level semantic memory or stronger distractor-aware matching could better distinguish visually similar UAVs. Tiny and blurred targets may benefit from super-resolution-assisted feature enhancement, uncertainty-aware box prediction, or localization heads designed for small aerial objects. Beyond visual tracking, integrating SATATrack with multi-modal sensing and closed-loop UAV control is an important step toward real-world anti-UAV systems, where tracking confidence should interact with path planning, camera control, and interception decisions.

%% file: section/6_conclusion.tex
In this paper, we study UAV anti-UAV tracking, a challenging air-to-air tracking scenario characterized by dual-dynamic motion, rapid background changes, tiny UAV targets, and severe target-distractor ambiguity. We propose \emph{SATATrack}, a Semantic-Aware Temporal Adaptation framework that integrates language-guided temporal modeling with test-time feature distribution alignment. Specifically, SACP injects semantic descriptions into state-space temporal updates and propagates target-aware context across backbone stages; TADA adapts normalization statistics at test time via temporal smoothing and train-test statistic blending without updating model parameters; and an auxiliary contrastive regularizer mines semantically confusing background patches as hard negatives during training. Extensive experiments validate the effectiveness of SATATrack in terms of performance, generalization, efficiency, and robustness under diverse UAV tracking conditions.

%% file: reference.bib
@article{li2024positive,
  author={Li, Xuelong},
  journal={IEEE TNNLS},
  title={Positive-Incentive Noise},
  year={2024},
  volume={35},
  number={6},
  pages={8708--8714},
  doi={10.1109/TNNLS.2022.3224577},
}

@article{zhang2025variational,
  author={Zhang, Hongyuan and Huang, Sida and Guo, Yubin and Li, Xuelong},
  journal={IEEE TPAMI},
  title={Variational Positive-Incentive Noise: How Noise Benefits Models},
  year={2025},
  volume={47},
  number={9},
  pages={8313--8320},
  doi={10.1109/TPAMI.2025.3575295},
}

@article{chen2026gvc,
  author={Chen, Xiangyu and Luo, Jixiang and Xu, Jingyu and Yi, Fangqiu and Zhang, Chi and Li, Xuelong},
  journal={Vicinagearth},
  title={Generative Video Compression: Towards 0.01\% Compression Rate for Video Transmission},
  year={2026},
  volume={3},
  number={1},
  pages={7},
  doi={10.1007/s44336-026-00035-2},
}

@article{an2026aiflow,
  author={An, Hongjun and Hu, Wenhan and Huang, Sida and Huang, Siqi and Li, Ruanjun and Liang, Yuanzhi and Shao, Jiawei and Song, Yiliang and Wang, Zihan and Yuan, Cheng and Zhang, Chi and Zhang, Hongyuan and Zhuang, Wenhao and Li, Xuelong},
  journal={Vicinagearth},
  title={{AI} Flow: Perspectives, Scenarios, and Approaches},
  year={2026},
  volume={3},
  number={1},
  pages={1},
  doi={10.1007/s44336-025-00031-y},
}

@article{huang2019got,
  author={Huang, Lianghua and Zhao, Xin and Huang, Kaiqi},
  journal={IEEE TPAMI}, 
  title={GOT-10k: A Large High-Diversity Benchmark for Generic Object Tracking in the Wild}, 
  year={2019},
  volume={43},
  number={5},
  pages={1562--1577},
}

@inproceedings{fan2019lasot,
  title={Lasot: A high-quality benchmark for large-scale single object tracking},
  author={Fan, Heng and Lin, Liting and Yang, Fan and others},
  booktitle={CVPR},
  pages={5374--5383},
  year={2019}
}

@inproceedings{muller2018trackingnet,
  title={Trackingnet: A large-scale dataset and benchmark for object tracking in the wild},
  author={Muller, Matthias and Bibi, Adel and Giancola, Silvio and Alsubaihi, Salman and Ghanem, Bernard},
  booktitle={ECCV},
  pages={300--317},
  year={2018}
}

@inproceedings{cao2021hift,
  title={HiFT: Hierarchical Feature Transformer for Aerial Tracking},
  author={Cao, Ziang and Fu, Changhong and Ye, Junjie and Li, Bowen and Li, Yiming},
  booktitle={ICCV},
  pages={15457--15466},
  year={2021}
}

@inproceedings{guo2021graph,
  title={Graph attention tracking},
  author={Guo, Dongyan and Shao, Yanyan and Cui, Ying and Wang, Zhenhua and Zhang, Liyan and Shen, Chunhua},
  booktitle={CVPR},
  pages={9543--9552},
  year={2021},
}

@inproceedings{yan2021lighttrack,
  title={LightTrack: Finding Lightweight Neural Networks for Object Tracking via One-Shot Architecture Search},
  author={Yan, Bin and Peng, Houwen and Wu, Kan and Wang, Dong and Fu, Jianlong and Lu, Huchuan},
  booktitle={CVPR},
  pages={15180--15189},
  year={2021},
}

@inproceedings{guo2020siamcar,
  title={SiamCAR: Siamese fully convolutional classification and regression for visual tracking},
  author={Guo, Dongyan and Wang, Jun and others},
  booktitle={CVPR},
  pages={6269--6277},
  year={2020},
}

@inproceedings{XuWLYY20,
  title={Siamfc++: Towards robust and accurate visual tracking with target estimation guidelines},
  author={Xu, Yinda and Wang, Zeyu and Li, Zuoxin and Yuan, Ye and Yu, Gang},
  booktitle={AAAI},
  volume={34},
  number={07},
  pages={12549--12556},
  year={2020},
}

@inproceedings{ChenZLZJ20,
  author    = {Zedu Chen and
               Bineng Zhong and
               Guorong Li and
               Shengping Zhang and
               Rongrong Ji},
  title     = {Siamese Box Adaptive Network for Visual Tracking},
  booktitle = {CVPR},
  pages     = {6667--6676},
  year      = {2020},
}

@article{jiang2021anti,
  title={Anti-UAV: A large-scale benchmark for vision-based UAV tracking},
  author={Jiang, Nan and Wang, Kuiran and Peng, Xiaoke and Yu, Xuehui and Wang, Qiang and Xing, Junliang and Li, Guorong and Guo, Guodong and Ye, Qixiang and Jiao, Jianbin and others},
  journal={IEEE TMM},
  volume={25},
  pages={486--500},
  year={2021}
}

@article{zhao2022vision,
  title={Vision-based anti-uav detection and tracking},
  author={Zhao, Jie and Zhang, Jingshu and Li, Dongdong and Wang, Dong},
  journal={IEEE TITS},
  volume={23},
  number={12},
  pages={25323--25334},
  year={2022}
}

@inproceedings{zhang2021learn,
  title={Learn to match: Automatic matching network design for visual tracking},
  author={Zhang, Zhipeng and Liu, Yihao and Wang, Xiao and others},
  booktitle={ICCV},
  pages={13339--13348},
  year={2021}
}

@inproceedings{lin2014microsoft,
  title={Microsoft coco: Common objects in context},
  author={Lin, Tsung-Yi and Maire, Michael and Belongie, Serge and others},
  booktitle={ECCV},
  pages={740--755},
  year={2014},
}

@article{guo2022divert,
  title={Divert More Attention to Vision-Language Tracking},
  author={Guo, Mingzhe and Zhang, Zhipeng and Fan, Heng and Jing, Liping},
  journal={NeurIPS},
  year={2022}
}

@inproceedings{zaveri2025improving,
  title={Improving accuracy and generalization for efficient visual tracking},
  author={Zaveri, Ram and Patel, Shivang and Gu, Yu and Doretto, Gianfranco},
  booktitle={Proceedings of the Winter Conference on Applications of Computer Vision},
  pages={9450--9460},
  year={2025}
}

@inproceedings{li2025mambalct,
  title={Mambalct: Boosting tracking via long-term context state space model},
  author={Li, Xiaohai and Zhong, Bineng and Liang, Qihua and Li, Guorong and Mo, Zhiyi and Song, Shuxiang},
  booktitle={Proceedings of the AAAI Conference on Artificial Intelligence},
  volume={39},
  number={5},
  pages={4986--4994},
  year={2025}
}

@inproceedings{wu2024mambanut,
  title={MambaNUT: Nighttime UAV Tracking via Mamba-based Adaptive Curriculum Learning},
  author={Wu, You and Yang, Xiangyang and Wang, Xucheng and Ye, Hengzhou and Zeng, Dan and Li, Shuiwang},
  booktitle={2025 IEEE/RSJ International Conference on Intelligent Robots and Systems (IROS)},
  pages={18864--18871},
  year={2025},
  organization={IEEE}
}

@inproceedings{wu2025learning,
  title={Learning Occlusion-Robust Vision Transformers for Real-Time UAV Tracking},
  author={Wu, You and Wang, Xucheng and Yang, Xiangyang and Liu, Mengyuan and Zeng, Dan and Ye, Hengzhou and Li, Shuiwang},
  booktitle={Proceedings of the Computer Vision and Pattern Recognition Conference},
  pages={17103--17113},
  year={2025}
}

@inproceedings{kang2025exploring,
  title={Exploring enhanced contextual information for video-level object tracking},
  author={Kang, Ben and Chen, Xin and Lai, Simiao and Liu, Yang and Liu, Yi and Wang, Dong},
  booktitle={Proceedings of the AAAI Conference on Artificial Intelligence},
  volume={39},
  number={4},
  pages={4194--4202},
  year={2025}
}

@article{gu2023mamba,
  title={Mamba: Linear-time sequence modeling with selective state spaces},
  author={Gu, Albert and Dao, Tri},
  journal={arXiv preprint arXiv:2312.00752},
  year={2023}
}

@inproceedings{zheng2024odtrack,
  title={Odtrack: Online dense temporal token learning for visual tracking},
  author={Zheng, Yaozong and Zhong, Bineng and Liang, Qihua and Mo, Zhiyi and Zhang, Shengping and Li, Xianxian},
  booktitle={Proceedings of the AAAI conference on artificial intelligence},
  volume={38},
  number={7},
  pages={7588--7596},
  year={2024}
}

@inproceedings{zhang2023all,
  title={All in One: Exploring Unified Vision-Language Tracking with Multi-Modal Alignment},
  author={Zhang, Chunhui and Sun, Xin and Yang, Yiqian and Liu, Li and Liu, Qiong and Zhou, Xi and Wang, Yanfeng},
  booktitle={Proceedings of the 31st ACM International Conference on Multimedia},
  pages={5552--5561},
  year={2023}
}

@article{ma2024unifying,
  title={Unifying Visual and Vision-Language Tracking via Contrastive Learning},
  author={Ma, Yinchao and Tang, Yuyang and Yang, Wenfei and Zhang, Tianzhu and Zhang, Jinpeng and Kang, Mengxue},
  journal={arXiv preprint arXiv:2401.11228},
  year={2024}
}

@inproceedings{li2023citetracker,
  title={CiteTracker: Correlating Image and Text for Visual Tracking},
  author={Li, Xin and Huang, Yuqing and He, Zhenyu and Wang, Yaowei and Lu, Huchuan and Yang, Ming-Hsuan},
  booktitle={Proceedings of the IEEE/CVF International Conference on Computer Vision},
  pages={9974--9983},
  year={2023}
}

@inproceedings{cao2022tctrack,
  title={Tctrack: Temporal contexts for aerial tracking},
  author={Cao, Ziang and Huang, Ziyuan and Pan, Liang and Zhang, Shiwei and Liu, Ziwei and Fu, Changhong},
  booktitle={Proceedings of the IEEE/CVF Conference on Computer Vision and Pattern Recognition},
  pages={14798--14808},
  year={2022}
}

@inproceedings{radford2021learning,
  title={Learning transferable visual models from natural language supervision},
  author={Radford, Alec and Kim, Jong Wook and Hallacy, Chris and \etal},
  booktitle={International Conference on Machine Learning},
  pages={8748--8763},
  year={2021},
}

@inproceedings{rezatofighi2019generalized,
  title={Generalized intersection over union: A metric and a loss for bounding box regression},
  author={Rezatofighi, Hamid and Tsoi, Nathan and Gwak, JunYoung and Sadeghian, Amir and Reid, Ian and Savarese, Silvio},
  booktitle={Proceedings of the IEEE/CVF Conference on Computer Vision and Pattern Recognition},
  pages={658--666},
  year={2019}
}

@inproceedings{li2019siamrpn++,
  title={Siamrpn++: Evolution of siamese visual tracking with very deep networks},
  author={Li, Bo and Wu, Wei and Wang, Qiang and Zhang, Fangyi and Xing, Junliang and Yan, Junjie},
  booktitle={Proceedings of the IEEE/CVF Conference on Computer Vision and Pattern Recognition},
  pages={4282--4291},
  year={2019}
}

@inproceedings{ye2022joint,
  title={Joint feature learning and relation modeling for tracking: A one-stream framework},
  author={Ye, Botao and Chang, Hong and Ma, Bingpeng and Shan, Shiguang and Chen, Xilin},
  booktitle={European Conference on Computer Vision},
  pages={341--357},
  year={2022},
}

@article{zhou2023joint,
  title={Joint Visual Grounding and Tracking with Natural Language Specification},
  author={Zhou, Li and Zhou, Zikun and Mao, Kaige and He, Zhenyu},
  journal={Proceedings of the IEEE/CVF Conference on Computer Vision and Pattern Recognition},
  pages={23151-23160},
  year={2023}
}

@inproceedings{xue2025similarity,
  title={Similarity-guided layer-adaptive vision transformer for UAV tracking},
  author={Xue, Chaocan and Zhong, Bineng and Liang, Qihua and Zheng, Yaozong and Li, Ning and Xue, Yuanliang and Song, Shuxiang},
  booktitle={Proceedings of the Computer Vision and Pattern Recognition Conference},
  pages={6730--6740},
  year={2025}
}

@inproceedings{wang2019fast,
  title={Fast online object tracking and segmentation: A unifying approach},
  author={Wang, Qiang and Zhang, Li and Bertinetto, Luca and Hu, Weiming and Torr, Philip HS},
  booktitle={CVPR},
  pages={1328--1338},
  year={2019}
}

@inproceedings{li2023adaptive,
  title={Adaptive and background-aware vision transformer for real-time uav tracking},
  author={Li, Shuiwang and Yang, Yangxiang and Zeng, Dan and Wang, Xucheng},
  booktitle={ICCV},
  pages={13989--14000},
  year={2023}
}

@inproceedings{feng2025atctrack,
  title={ATCTrack: Aligning Target-Context Cues with Dynamic Target States for Robust Vision-Language Tracking},
  author={Feng, Xiaokun and Hu, Shiyu and Li, Xuchen and Zhang, Dailing and Wu, Meiqi and Zhang, Jing and Chen, Xiaotang and Huang, Kaiqi},
  booktitle={ICCV},
  pages={19850--19861},
  year={2025}
}

@inproceedings{li2025dynamic,
  title={Dynamic Updates for Language Adaptation in Visual-Language Tracking},
  author={Li, Xiaohai and Zhong, Bineng and Liang, Qihua and Mo, Zhiyi and Nong, Jian and Song, Shuxiang},
  booktitle={CVPR},
  pages={19165--19174},
  year={2025}
}

@inproceedings{chen2025sutrack,
  title={Sutrack: Towards simple and unified single object tracking},
  author={Chen, Xin and Kang, Ben and Geng, Wanting and Zhu, Jiawen and Liu, Yi and Wang, Dong and Lu, Huchuan},
  booktitle={AAAI},
  volume={39},
  number={2},
  pages={2239--2247},
  year={2025}
}

@inproceedings{zhang2025mambatrack,
  title={Mambatrack: Exploiting dual-enhancement for night uav tracking},
  author={Zhang, Chunhui and Liu, Li and Wen, Hao and Zhou, Xi and Wang, Yanfeng},
  booktitle={ICASSPP},
  pages={1--5},
  year={2025},
  organization={IEEE}
}

@inproceedings{mueller2016benchmark,
  title={A Benchmark and Simulator for {UAV} Tracking},
  author={Mueller, Matthias and Smith, Neil and Ghanem, Bernard},
  booktitle={ECCV},
  pages={445--461},
  year={2016}
}

@inproceedings{du2018unmanned,
  title={The unmanned aerial vehicle benchmark: Object detection and tracking},
  author={Yu, Hongyang and Li, Guorong and Zhang, Weigang and others},
  booktitle={ECCV},
  pages={370--386},
  year={2018}
}

@ARTICLE{zhang2022webuav,
  author={Zhang, Chunhui and Huang, Guanjie and Liu, Li and Huang, Shan and Yang, Yinan and Wan, Xiang and Ge, Shiming and Tao, Dacheng},
  journal={IEEE TPAMI}, 
  title={WebUAV-3M: A Benchmark for Unveiling the Power of Million-Scale Deep UAV Tracking}, 
  year={2023},
  volume={45},
  number={7},
  pages={9186-9205},
  doi={10.1109/TPAMI.2022.3232854}
}

@article{zhu2020vision,
  title={Detection and tracking meet drones challenge},
  author={Zhu, Pengfei and Wen, Longyin and Du, Dawei and Bian, Xiao and Fan, Heng and Hu, Qinghua and Ling, Haibin},
  journal={IEEE Transactions on Pattern Analysis and Machine Intelligence},
  volume={44},
  number={11},
  pages={7380--7399},
  year={2021}
}

@article{huang2023anti,
  title={Anti-UAV410: A Thermal Infrared Benchmark and Customized Scheme for Tracking Drones in the Wild},
  author={Huang, Bo and Li, Jianan and Chen, Junjie and Wang, Gang and Zhao, Jian and Xu, Tingfa},
  journal={IEEE Transactions on Pattern Analysis and Machine Intelligence},
  year={2023},
  publisher={IEEE}
}

@article{zhu2023evidential,
  title={Evidential detection and tracking collaboration: New problem, benchmark and algorithm for robust anti-uav system},
  author={Zhu, Xue-Feng and Xu, Tianyang and Zhao, Jian and Liu, Jia-Wei and Wang, Kai and Wang, Gang and Li, Jianan and Wang, Qiang and Jin, Lei and Zhu, Zheng and others},
  journal={arXiv preprint arXiv:2306.15767},
  year={2023}
}

@article{xu2025tri,
  title={A Tri-Modal Dataset and a Baseline System for Tracking Unmanned Aerial Vehicles},
  author={Xu, Tianyang and Gu, Jinjie and Zhu, Xuefeng and Wu, XiaoJun and Kittler, Josef},
  journal={arXiv preprint arXiv:2511.18344},
  year={2025}
}

@inproceedings{dong2025securing,
  title={Securing the Skies: A Comprehensive Survey on Anti-UAV Methods, Benchmarking, and Future Directions},
  author={Dong, Yifei and Wu, Fengyi and Zhang, Sanjian and Chen, Guangyu and Hu, Yuzhi and Yano, Masumi and Sun, Jingdong and Huang, Siyu and Liu, Feng and Dai, Qi and others},
  booktitle={Proceedings of the Computer Vision and Pattern Recognition Conference},
  pages={6659--6673},
  year={2025}
}

@inproceedings{wang2021towards,
  title={Towards more flexible and accurate object tracking with natural language: Algorithms and benchmark},
  author={Wang, Xiao and Shu, Xiujun and Zhang, Zhipeng and Jiang, Bo and Wang, Yaowei and Tian, Yonghong and Wu, Feng},
  booktitle={IEEE CVPR},
  pages={13763--13773},
  year={2021}
}

@article{devlin2018bert,
  title={BERT: Pre-training of Deep Bidirectional Transformers for Language Understanding},
  author={Devlin, Jacob and Chang, Ming-Wei and Lee, Kenton and Toutanova, Kristina},
  journal={arXiv preprint arXiv:1810.04805},
  year={2018}
}

@inproceedings{bertinetto2016fully,
  title={Fully-convolutional siamese networks for object tracking},
  author={Bertinetto, Luca and Valmadre, Jack and Henriques, Joao F and others},
  booktitle={ECCV},
  pages={850--865},
  year={2016},
}

@inproceedings{danelljan2017eco,
  title={Eco: Efficient convolution operators for tracking},
  author={Danelljan, Martin and Bhat, Goutam and Shahbaz Khan, Fahad and Felsberg, Michael},
  booktitle={CVPR},
  pages={6638--6646},
  year={2017},
}

@inproceedings{danelljan2019atom,
  title={Atom: Accurate tracking by overlap maximization},
  author={Danelljan, Martin and Bhat, Goutam and Khan, Fahad Shahbaz and Felsberg, Michael},
  booktitle={Proceedings of the IEEE/CVF Conference on Computer Vision and Pattern Recognition},
  pages={4660--4669},
  year={2019}
}

@inproceedings{dosovitskiy2020image,
  title={An Image is Worth 16x16 Words: Transformers for Image Recognition at Scale},
  author={Dosovitskiy, Alexey and Beyer, Lucas and Kolesnikov, Alexander and \etal},
  booktitle={International Conference on Learning Representations},
  year={2020}
}

@inproceedings{chen2021transformer,
  title={Transformer tracking},
  author={Chen, Xin and Yan, Bin and Zhu, Jiawen and Wang, Dong and Yang, Xiaoyun and Lu, Huchuan},
  booktitle={CVPR},
  pages={8126--8135},
  year={2021},
}

@inproceedings{wang2021transformer,
  title={Transformer Meets Tracker: Exploiting Temporal Context for Robust Visual Tracking},
  author={Wang, Ning and Zhou, Wengang and others},
  booktitle={CVPR},
  pages={1571--1580},
  year={2021},
}

@inproceedings{yan2021learning,
  title={Learning spatio-temporal transformer for visual tracking},
  author={Yan, Bin and Peng, Houwen and Fu, Jianlong and Wang, Dong and Lu, Huchuan},
  booktitle={Proceedings of the IEEE/CVF Conference on Computer Vision and Pattern Recognition},
  pages={10448--10457},
  year={2021}
}

@article{zhu2024vision,
  title={Vision mamba: Efficient visual representation learning with bidirectional state space model},
  author={Zhu, Lianghui and Liao, Bencheng and Zhang, Qian and Wang, Xinlong and Liu, Wenyu and Wang, Xinggang},
  journal={arXiv preprint arXiv:2401.09417},
  year={2024}
}

@article{zhang2025far,
  title={How Far are Modern Trackers from UAV-Anti-UAV? A Million-Scale Benchmark and New Baseline},
  author={Zhang, Chunhui and Liu, Li and Zhang, Zhipeng and Wang, Yong and Wen, Hao and Zhou, Xi and Ge, Shiming and Wang, Yanfeng},
  journal={arXiv preprint arXiv:2512.07385},
  year={2025}
}

@article{tian2024fast,
  title={Fast-iTPN: Integrally pre-trained transformer pyramid network with token migration},
  author={Tian, Yunjie and Xie, Lingxi and Qiu, Jihao and Jiao, Jianbin and Wang, Yaowei and Tian, Qi and Ye, Qixiang},
  journal={IEEE Transactions on Pattern Analysis and Machine Intelligence},
  volume={46},
  number={12},
  pages={9766--9779},
  year={2024},
  publisher={IEEE}
}

@inproceedings{shao2025pura,
  title={PURA: Parameter Update-Recovery Test-Time Adaption for RGB-T Tracking},
  author={Shao, Zekai and Hu, Yufan and Fan, Bin and Liu, Hongmin},
  booktitle={Proceedings of the Computer Vision and Pattern Recognition Conference},
  pages={22089--22098},
  year={2025}
}

@article{qiao2026bidirectional,
  title={Bidirectional prototype-reward co-evolution for test-time adaptation of vision-language models},
  author={Qiao, Xiaozhen and Huang, Peng and Yuan, Jiakang and Guo, Xianda and Ye, Bowen and Xue, Chaocan and Zheng, Ye and Sun, Zhe and Li, Xuelong},
  journal={IEEE Transactions on Multimedia},
  year={2026},
  publisher={IEEE}
}

@article{qiao2025class,
  title={Class-Aware Prototype Learning with Negative Contrast for Test-Time Adaptation of Vision-Language Models},
  author={Qiao, Xiaozhen and Zhao, Jingkai and Jiang, Yuqiu and Guo, Xianda and Sun, Zhe and Zhang, Hongyuan and Li, Xuelong},
  journal={arXiv preprint arXiv:2510.19802},
  year={2025}
}
